\documentclass{article}

\PassOptionsToPackage{numbers, compress}{natbib}

\usepackage[preprint]{neurips_2025}

\usepackage[utf8]{inputenc} % allow utf-8 input
\usepackage[T1]{fontenc}    % use 8-bit T1 fonts
\usepackage{hyperref}       % hyperlinks
\usepackage{url}            % simple URL typesetting
\usepackage{booktabs}       % professional-quality tables
\usepackage{amsfonts}       % blackboard math symbols
\usepackage{nicefrac}       % compact symbols for 1/2, etc.
\usepackage{microtype}      % microtypography
\usepackage{xcolor}         % colors
\usepackage[pdftex]{graphicx}

\usepackage{subcaption}
\usepackage{enumitem}
\usepackage{wrapfig}

\usepackage{amsmath}
\usepackage{bm}
\usepackage{mathtools}
\usepackage{amssymb}
\usepackage{amsthm}

\usepackage{pifont}

\usepackage{tikz}
\usetikzlibrary{plotmarks}
\usetikzlibrary{3d}
\usetikzlibrary{shapes, arrows, fit, calc, positioning, matrix}
\usetikzlibrary{decorations.markings}

\title{Knowledge Insulating Vision-Language-Action Models: Train Fast, Run Fast, Generalize Better}

\author{%
  Danny Driess
  ~~~~~Jost Tobias Springenberg
  ~~~~~Brian Ichter
  ~~~~~Lili Yu \\[0.1cm]
  \textbf{Adrian Li-Bell}
  ~~~~~\textbf{Karl Pertsch}
  ~~~~~\textbf{Allen Z.\ Ren}
  ~~~~~\textbf{Homer Walke}\\[0.1cm]
  \textbf{Quan Vuong}
  ~~~~~\textbf{Lucy Xiaoyang Shi}
  ~~~~~\textbf{Sergey Levine}\\[0.2cm]
  Physical Intelligence
}

\begin{document}

\maketitle

\begin{abstract}
Vision-language-action (VLA) models provide a powerful approach to training control policies for physical systems, such as robots, by combining end-to-end learning with transfer of semantic knowledge from web-scale vision-language model (VLM) training. However, the constraints of real-time control are often at odds with the design of VLMs: the most powerful VLMs have tens or hundreds of billions of parameters, presenting an obstacle to real-time inference, and operate on discrete tokens rather than the continuous-valued outputs that are required for controlling robots. To address this challenge, recent VLA models have used specialized modules for efficient continuous control, such as action experts or continuous output heads, which typically require adding new untrained parameters to the pretrained VLM backbone. While these modules improve real-time and control capabilities, it remains an open question whether they preserve or degrade the semantic knowledge contained in the pretrained VLM, and what effect they have on the VLA training dynamics. In this paper, we study this question in the context of VLAs that include a continuous diffusion or flow matching action expert, showing that naively including such experts significantly harms both training speed and knowledge transfer. We provide an extensive analysis of various design choices, their impact on performance and knowledge transfer, and propose a technique for insulating the VLM backbone during VLA training that mitigates this issue.
Videos are available at \href{https://pi.website/research/knowledge_insulation}{https://pi.website/research/knowledge\_insulation}.
\end{abstract}

\section{Introduction}

The success of large language models (LLMs) can be attributed to the availability of large-scale datasets combined with powerful model architectures such as transformers that are trained with a next-token prediction objective on trillions of tokens. LLMs can be prompted to solve all sorts of tasks, from writing poems and code to solving competition-level math problems, and can further be adapted to solve visual reasoning problems when extended with multi-modal encoders to produce vision-language models (VLMs). A natural next step to bring the power of LLMs to the physical world is to further extend them to take physical actions, resulting in vision-language action (VLA) models that can control robots to follow language commands, combining the power of end-to-end robotic learning with the semantic knowledge distilled from web-scale vision-language pretraining~\cite{rt22023arxiv, kim2024openvla, black2024pi_0}.
However, adapting LLMs and VLMs to real-world control requires addressing a number of new challenges. Most physical systems (e.g., robots) require continuous and precise commands, such as joint angles or target poses, that must be produced in real time at a high frequency. Autoregressive decoding of discrete tokens is poorly suited to this kind of high-frequency continuous control, both because of the limited resolution of discretized actions and because of the computational cost of autoregressive decoding with large models, a challenge only exacerbated by ever larger models. Furthermore, physical systems typically produce more complex observations than VLMs are trained for, such as multi-view images and proprioceptive states. These differences necessitate modifications to the original VLM architecture to accommodate robotic control.

Consequently, the robotics community has developed architectures that are particularly well-suited to the demands of real-time continuous control~\citep{zhao2023learning,chi2023diffusion,team2024octo,zhao2024alohaunleashedsimplerecipe,black2024pi_0,liu2025hybridvla,bjorck2025gr00t,bu2025agibot,kim2025fine,pi2025pi05}. While a number of different designs have been successful, a common theme is that models adapted for effective dexterous control typically augment a transformer or VLM backbone with some sort of adapter for continuous inputs and outputs, with the latter most often utilizing, for example, diffusion or flow matching with action chunks (short sequences of future actions)~\cite{zhao2023learning}. This enables the model to represent complex continuous action distributions, select very precise actions, and capture dexterous high-frequency skills. However, when these additional modules are added to a pre-trained VLM to create VLAs, they typically need to be initialized from scratch, and the VLA training process must ``graft'' them onto the VLM backbone. This raises an important question: \emph{how much do VLAs augmented with these continuous state and action adapters actually inherit and benefit from web-scale pre-training?}

\begin{figure}
    \centering
    \vspace{-2mm}
    \includegraphics[width=\linewidth]{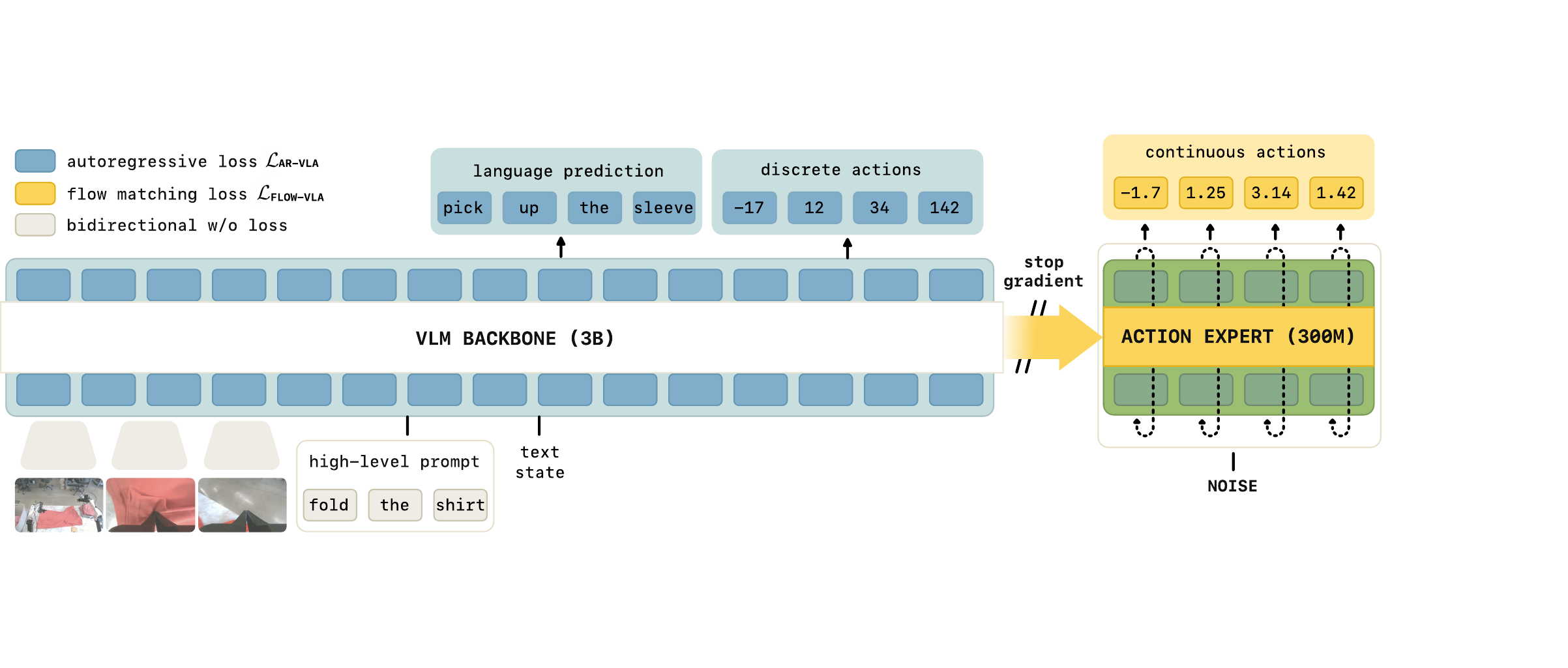}
    \vspace{-5mm}
    \caption{\small The key idea of our approach is to train the VLM backbone with a next-token prediction loss on discretized actions and general VLM data to learn good representations, while the action expert is trained with flow-matching on continuous actions. Gradients do not flow from the action expert to the backbone, insulating the knowledge of the backbone. At inference time, generating continuous actions with the smaller action expert is desirable for fast and precise control, while representation learning with discrete actions and general VLM data makes the model train fast and leads to better generalization by transferring knowledge from VLM data into robot actions. As experiments show, having both action representations at training time is crucial.}
    \label{fig:model}
    \vspace{-6mm}
\end{figure}

In this work, we observe that prior approaches for finetuning VLMs with continuous outputs can, perhaps unsurprisingly, lead to significantly worse training dynamics, as they rely on gradients from continuous adapters (e.g. diffusion heads) for the training signal. This can degrade both their ability to interpret language commands and the overall performance of the resulting VLA policy. To address this challenge, we propose a training recipe that addresses these issues, which we refer to as \emph{knowledge insulation}. The key idea behind knowledge insulation is to fine-tune the VLM backbone with discretized actions while \emph{simultaneously} adapting an action expert to produce continuous actions (e.g., via flow matching or diffusion) \emph{without} propagating its gradients back into the VLM backbone. We illustrate this in Figure~\ref{fig:model}. In effect, the discrete action tokens provide a substitute learning signal that is unaffected by the uninitialized weights of the action expert, such that the VLM still learns appropriate representations for robotic control, but without the disruption that would stem from gradients from the action expert. This approach has additional advantages: first, using next-token prediction makes the model learn much faster and more stably. Second, using an action expert still enables fast inference. Third, our recipe enables us to co-train a model on general vision-language data, bringing the advantages of VLAs back into our model.
Our experimental evaluation provides an extensive analysis of the various modeling choices in continuous-action VLAs, building on the $\pi_0$ model architecture~\cite{black2024pi_0}.
We evaluate on complex, long-horizon robotic manipulation tasks, including mobile bimanual robots, as well as open-source benchmarks such as DROID and LIBERO.

\section{Related work}

\textbf{Multi-modal large language models.}
In this work, we investigate how to integrate \emph{robot actions} as a new modality into pre-trained VLMs. In the literature, a common technique is to embed new input modalities into discrete or continuous (``soft'') tokens~\cite{liu2023llava, laurencon2023obelics, chen2022visualgpt, team2024chameleon}. Following earlier works on multi-modal cross-attention \cite{alayrac2022flamingo}, recent work showed that for multi-modal \emph{generative} modeling, e.g., interleaved image, text and speech prediction, separating modalities into modality-specific ``expert'' networks that cross-attend to each other can prevent interference and lead to higher quality predictions \cite{liang2024mixture,shi2024xi,cogvlm}. While these works trained vision-language-speech models, we are interested in using similar architectures for fusing a new modality, robot actions, into pre-trained VLMs.

\textbf{Vision-language-action models (VLAs).}
Vision-language-action models have recently been proposed as a promising approach for generalizable robot control~\citep{driess2023palm, rt22023arxiv, kim2024openvla, black2024pi_0, wen2024tinyvlafastdataefficientvisionlanguageaction, zhen20243dvla, liu2024rdt, li2024cogact, belkhale2024minivla, szot2024multimodal, pertsch2025fast, geminirobotics2025, wen2025dexvla, bjorck2025gr00t, huang2025otter}. The core idea of VLAs is to fine-tune pre-trained vision-language models (VLMs) for action prediction. Such VLAs scale favorably to large-scale robot datasets~\citep{dasari2019robonet, ebert2021bridge, walke2023bridgedata, open_x_embodiment_rt_x_2023, khazatsky2024droid, bharadhwaj2023roboagent, fang2024rh20t, shafiullah2023bringingrobotshome, contributors2025agibotworld}, and have been shown to transfer knowledge from web-scale VLM pre-training to improve policy generalization~\citep{rt22023arxiv, kim2024openvla}. To enable VLM fine-tuning on action data, VLA training pipelines typically map continuous actions to a sequence of discrete action \emph{tokens}, either through simple binning discretization~\citep{rt22023arxiv, kim2024openvla}, or more advanced, compression-based tokenization approaches~\citep{pertsch2025fast}. The VLA is then trained via standard autoregressive next-token prediction. While this strategy is effective on simpler, low-frequency control tasks, it has two drawbacks: (1)~the mapping from continuous to discrete action tokens can be lossy, and (2)~decoding actions autoregressively leads to slow policy inference~\citep{pertsch2025fast, belkhale2024minivla}. As a result, modern autoregressive VLAs typically run at control frequencies lower than 2Hz~\citep{pertsch2025fast}, making them impractical for many high-frequency tasks.

\textbf{Fast and continuous action decoding mechanisms in VLAs.}
To address these concerns, multiple prior works have explored alternative action decoding mechanisms for VLAs that retain continuous action outputs and fast inference. These approaches typically introduce new weights and losses during VLA action fine-tuning,
often via diffusion prediction heads~\citep{liu2024rdt, wen2025dexvla, liu2025hybridvla, bjorck2025gr00t} or flow-matching based ``action experts'' \citep{black2024pi_0,bu2025agibot} that attend to features in the VLM backbone. While these approaches enable fast inference, simply adding new weights and training losses during fine-tuning comes with its own issues: such VLAs are often significantly slower to train than their autoregressive counter-parts, and suffer from reduced web data transfer~\citep{pertsch2025fast}.
\citet{liu2025hybridvla} introduce a \emph{hybrid} autoregressive-diffusion training approach
but still require slow autoregressive action decoding at inference time. Several works, including OpenVLA-OFT \citep{kim2025fine} and $\pi_{0.5}$~\citep{pi2025pi05}, employ a two-stage procedure, where the model is first trained with autoregressive discretization, and then fine-tuned to a target domain with continuous outputs.

\textbf{$\pi_0$ and $\pi_{0.5}$ models.}
We build on the $\pi_0$~\citep{black2024pi_0} and $\pi_0\text{-FAST}$~\citep{pertsch2025fast} VLAs. $\pi_0$ introduced a continuous action expert, which can capture complex continuous distributions over action chunks, allows for efficient inference, and enables continuous control of dexterous tasks, such as folding laundry. However, the $\pi_0$ recipe by itself, as we show in our experiments, leads to degradation in terms of both language following and training speed, as the gradients from the action expert degrade the pre-trained VLM backbone. $\pi_0\text{-FAST}$ addresses this by using tokenized actions, using a DCT-based tokenizer that allows for efficient discretization of complex action chunks, but at the cost of requiring expensive autoregressive inference and degrading the ability to perform delicate and dynamic tasks, as we also illustrate in our experiments.
$\pi_{0.5}$~\citep{pi2025pi05} first trains with only FAST tokenized actions, and then adds a randomly initialized action expert in post-training for fine-tuning on mobile manipulation data (by joint-training).
Our work formalizes the approach of $\pi_{0.5}$ and extends it to develop a single-stage training recipe, where the VLM backbone is adapted for robotic control with discrete tokens while the action expert is \emph{simultaneously} trained to produce continuous actions, providing the best of both worlds. We rigorously ablate different mechanisms for knowledge preservation and co-training in our experiments.
We thus propose the first VLA recipe that trains quickly, retains VLM knowledge, and supports high-frequency control with continuous action outputs.

\section{Standard vision-language-action (VLA) model training recipes}\label{sec:standardVLARecipes}
We describe standard recipes for building and training vision-language-action models (VLAs).
The idea in training a VLA $\pi$ is to adapt a vision-language model (VLM) to output robot actions $a\in\mathbb{R}^d$ conditioned on image observations $I_{1:V}$, the robot's proprioceptive state $q\in\mathbb{R}^s$, and a natural language instruction $\ell$ as input, i.e.\ $a \sim \pi(\cdot | I_{1:V}, q, \ell)$.
The promise of VLAs is to inherit knowledge of the underlying VLM pre-trained on internet-scale data when finetuning it to robot actions.

\textbf{Action representations.}
Robot actions $a\in\mathbb{R}^d$ are, in most cases, real-valued vectors that typically represent robot joint angles or end-effector coordinates.
A common strategy is to employ so-called action chunking \cite{zhao2023learning}, i.e.\ to predict a trajectory of robot actions $a_{1:H}$ relative to the current robot state.
To adapt a VLM to a VLA, there are multiple choices of how to represent those action chunks.

\textbf{Na\"{i}ve discretization.}
In the simplest case, each dimension of each action in a chunk is discretized, and then each discretization bin is associated with a special text token \cite{rt22023arxiv}.
This way, a chunk $a_{1:H}$ is mapped into $H \cdot d$ tokens. Robot action prediction then is framed as a next-token prediction problem and the model can be trained as if it was a non-robot specific VLM with a cross-entropy loss.

\textbf{Temporal action abstractions.}
The disadvantage of na\"{i}ve discretization is that for high-frequency and high-dim.\ systems the number of tokens to represent actions grows quickly, which greatly increases the computational cost and leads to slow training convergence. Recent work, e.g.\ PRISE \citep{zheng2024prise}, FAST \citep{pertsch2025fast}, mitigate this effect by applying a transformation that compresses information in time. We use FAST for encoding actions, which applies a discrete cosine transform to each dimension in the action chunk, followed by quantization and byte-pair encoding~\citep{gage1994new} to produce action tokens.

\textbf{Diffusion and flow matching.}
A number of recently proposed VLA models have used diffusion or flow matching~\cite{lipman2022flow, liu2022flow} to generate continuous actions, and our own experiments follow the design of $\pi_0$ in using a flow matching ``action expert''~\cite{black2024pi_0}, as shown in Figure~\ref{fig:model}.
For the flow matching time index $\tau\in[0,1]$, the input to the model is a noised version of the action chunk $a_{1:H}^{\tau, \omega} = \tau a_{1:H} + (1-\tau)\omega$, $\omega \sim \mathcal{N}(0, \mathbf{I})$, and the model is trained to predict the flow $\omega - a_{1:H}$.
At inference time, this flow field is integrated to denoise $\omega$ to the final action chunk.

\textbf{State representations.}
We consider three different representations for the robot's proprioceptive state, namely to represent it as text (``text state'') after discretization, to use special tokens (``special token state'') also with discretization, and by directly mapping the continuous state into the backbone with a learned projection (``continuous state'').
We refer to Sec.~\ref{sec:stateRepresentations} for more details and discussion.

\textbf{VLA architectures, training, \& mixture of experts.}
Most VLAs are built from a multimodal transformer, usually initialized with pre-trained VLM weights.
Here, we describe a general form of transformer-based VLA architectures.
Our model
\begin{align}
    \pi_\theta(y | I_{1:v}, q, \ell) = p_\theta(y_{1:n} | x_{1:n}) = p\Big( y_{1:n} \Big| f_\theta\Big(\big(\phi_{\rho(i)}(x_i)\big)_{i=1}^n, A\big((\rho(i))_{i=1}^n\big), (\rho(i))_{i=1}^n\Big)\Big) \label{eq:transformerModel}
\end{align}
maps a sequence of $n$ multimodal input tokens $x_i$ to probabilities over a sequence of $n$ multimodal output tokens $y$. For VLAs, typically $y = y^a$ corresponds to action targets. Previous work has considered training one model jointly for action prediction and VLM tasks (for which $y = y^\ell$ is a tokenized text output) \citep{driess2023palm,rt22023arxiv}.
As indicated by its modality type $\rho : i \mapsto \{\text{image}, \text{word}, \text{action}, \text{state}, \ldots\ldots\}$, each token can be a text token ($x_i^\ell\in\mathbb{N}$), an image patch ($x_i^I\in\mathbb{R}^{p\times p \times 3}$), or a continuous input ($x_i\in\mathbb{R}^d$) such as robot states or actions.
The tokens are embedded with different encoders $\phi_j : \mathcal{T}_j \rightarrow \mathbb{R}^{d_e}$, where $\mathcal{T}_j$ is the space of all multimodal tokens of type $j$, and $d_e$ the embedding dimension of the model.
Image patches are encoded with a vision-transformer, text tokens with an embedding matrix, and continuous inputs via an affine projection.
The attention mask $A\big((\rho(i))_{i=1}^n\big)\in\{-\infty, 0\}^{n\times n}$ indicates which tokens can attend to each other.

\looseness=-1
A transformer \cite{vaswani2017attention} is a function $f:\mathbb{R}^{n\times d_e} \rightarrow \mathbb{R}^{n\times d_e}$ that maps $n$ input embeddings to $n$ output embeddings.
It is built by stacking multiple blocks that themselves are composed of an attention layer, a feedforward layer, and normalization layers.
Let $X = x_{1:n}\in\mathbb{R}^{n\times d_e}$.
The attention layer in a standard transformer is computed as $\mathrm{attn}(X) = E(X)W_V$, $W_V\in\mathbb{R}^{d_e\times d_v}$, where $E(X) = P(X)V(X)$, $P(X) = \mathrm{softmax}(Q(X)K(X)^T)$, and $Q(\cdot)$, $K(\cdot)$, $V(\cdot)$ are the so-called query, key, and value projections, e.g.\ $Q(X) = XQ_m$, $Q_m\in\mathbb{R}^{d_e \times d_q}$ with $d_q$ being the dimension of the projection.
Compared to a standard transformer, our model processes different tokens with separate weights, as proposed in \cite{liang2024mixture}.
As $\pi_0$ \cite{black2024pi_0}, we initialize the VLM from PaliGemma \citep{beyer2024paligemma} and use a smaller set of weights for action tokens which significantly reduces the inference time when generating actions.
The backbone and action tokens have their own query, key and value projections, but the dimensions $d_q$, $d_k$, $d_v$ of those projections are the same such that experts can interact with each other.

Most VLAs are trained on large robot behavior cloning datasets.
For autoregressive architectures, the standard training procedure is to minimize the negative log-likelihood of target tokens
\vspace{-2mm}
\begin{equation}
    \mathcal{L}_\text{AR-VLA}(\theta) =\mathbb{E}_{(x, y) \sim \mathcal{D}}[- \log p_\theta(y_{1:n} | x_{1:n}) ] = \mathbb{E}_{(x, y) \sim \mathcal{D}} -  \sum_{j=1}^{n-1} M_j \log p_\theta(y_{j+1} | x_{1:{j}}),
    \label{eq:ar-vla}
\end{equation}
where $M$ is a loss mask indicating which tokens should be predicted and $\mathcal{D}$ is a dataset (and we typically assume $x = y$).
In cases where flow-matching is used for action prediction the loss is
\vspace{-2mm}
\begin{equation}
    \mathcal{L}_\text{FLOW-VLA}(\theta) = \mathbb{E}_{\mathcal{D}, \tau, \omega} \Big[ \left\|\omega - a_{1:H} - f^a(a^{\tau, \omega}_{1:H})\right\|^2 \Big].
    \label{eq:flow-vla}
\end{equation}

\section{Problems with standard VLA recipes}\label{sec:problemsWithVLAs}
\begin{wrapfigure}{r}{0.6\linewidth}
    \centering
    \vspace{-0.6cm}
    \includegraphics[width=\linewidth]{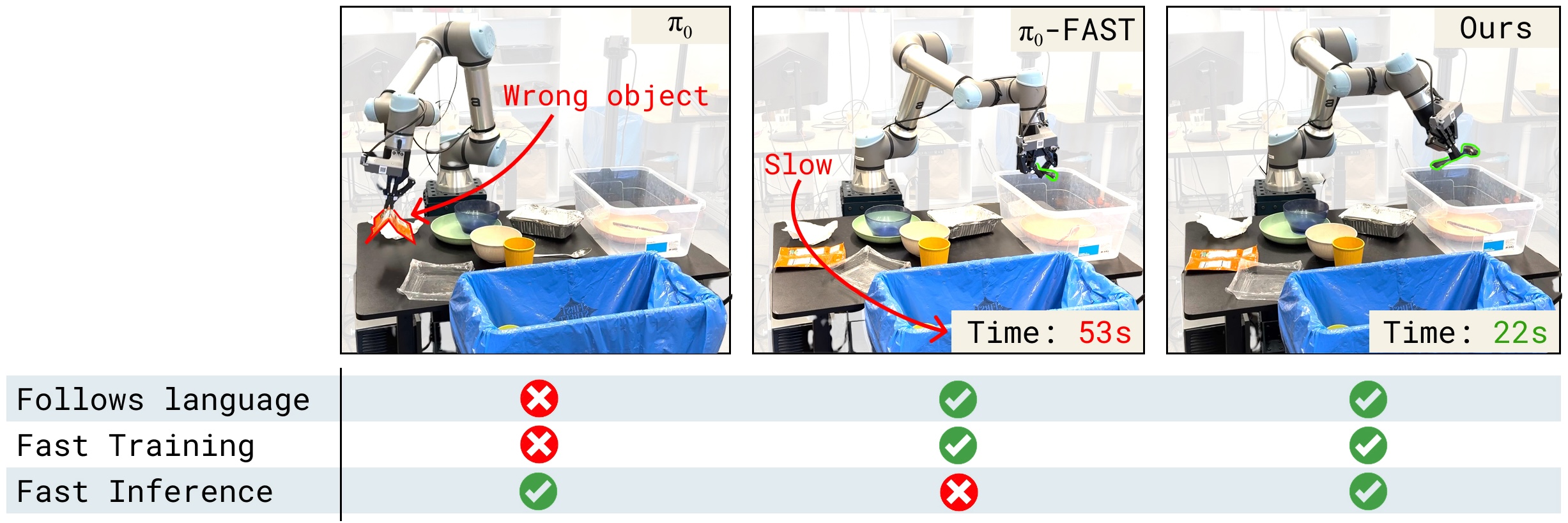}
    \vspace{-6mm}
    \caption{\small Problems with standard VLA recipes. The robot is instructed to bus the spoon into the bin. $\pi_0$~\citep{black2024pi_0} (left) ignores the command and grasps a piece of trash instead. $\pi_0$-FAST~\citep{pertsch2025fast} (middle) will eventually succeed but its inference time is very slow. Our recipe (right) solves the task, has fast inference, and the model converges very quickly to good performance (cf.\ Fig.~\ref{fig:UR5TrainingSpeed}).}
    % \vspace{-0.5cm}
    \label{fig:problemsWithStandardVLAs}
    \vspace{-0.5cm}
\end{wrapfigure}

In Fig.~\ref{fig:problemsWithStandardVLAs}, we visualize problems with current recipes for training VLAs.

\textbf{Autoregressive VLAs are slow.}
Autoregressive VLAs cast the problem of predicting real-valued actions as a discrete next-token prediction problem, which both limits the resolution of values the model can represent and results in slow, sequential inference.
The inference time of $\pi_0$-FAST for predicting a 1-second action chunk is $\approx$750~ms on an RTX4090 GPU \cite{pertsch2025fast}, which, as we show in the experiments, can lead to dynamics mismatches and slow overall trajectories.

\textbf{Robotic specific architectures and modality adapters don't benefit as much from VLM pre-training.}
Architectures like $\pi_0$~\citep{black2024pi_0} or GROOT~\citep{bjorck2025gr00t} contain robotics specific modules that enable faster inference. For example,
the action expert in the $\pi_0$ architecture has fewer parameters than the VLM backbone, and hence $\pi_0$ can achieve a control frequency of 10~Hz, which is much faster than autoregressive VLAs (1.3~Hz).
While parts of these models are initialized from pre-trained VLMs (e.g.\ the vision encoder or language model backbone), the robotics-specific modules are initialized from scratch. We show that naive training with such a randomly initialized action expert harms the models' ability to follow language commands (presumably due to gradient interference).

\textbf{VLM pretraining does not have sufficient representations for robotics---freezing doesn't work.}
Intuitively, the easiest way of maintaining the knowledge from VLM pretraining, and thus avoiding the problem mentioned above, would be to freeze the pre-trained weights and only train the newly added, robotics-specific weights. However, current VLMs are not pre-trained with robotics data. As a result, their representations, when frozen, are insufficient for training highly performant policies, as we show in our experiments, cf.\ Fig.~\ref{fig:arxSinglePerformance:performance} and Fig.~\ref{fig:shirtFolding} (0\% performance).

\section{Improving VLAs with co-training, joint-training \& knowledge insulation}
We consider a number of measures in order to overcome the limitations of prior VLA approaches outlined in Sec.~\ref{sec:problemsWithVLAs}. In particular, we propose:
\vspace{-0.2cm}
\begin{enumerate}[leftmargin=0.9cm]
    \itemsep0em
    \item to train the model on both autoregressive and flow-matching action prediction jointly at the same time (joint-training). The model uses the (smaller) action expert to produce continuous actions for fast inference at test time.
    The autoregressive objective is only used at training time as a representation learning objective, which enables the model to train much faster.
    \item to co-train the model on non-action datasets such as general vision-language data, and robot planning data (VLM data co-training). Training on these data sources ensure that the model loses less of its knowledge when adapting it to a VLA. 
    \item to stop the gradient flow between the action expert and the backbone weights. This way, when adapting the pre-trained VLM to a VLA, the newly initialized weights of the action expert don't interfere with the pre-trained weights.
\end{enumerate}

\subsection{Co-training \& representation learning with joint discrete/continous action prediction}\label{sec:co-training}
To enable effective co-training with VLM data, enhance knowledge transfer from language to policies, and allow for fast training, we consider combining autoregressive language and discrete action predictions as well as flow-matching modeling of continuous actions all \emph{in one model}. 
In particular, we propose to learn a model from which we can sample both, real-valued action chunks $a_{1:H}$, $a_i\in\mathbb{R}^d$, and text $\hat{\ell}$, i.e.\ the output space of our model is $y = (a_{1:H}, y^{\ell,a})$, where $a_{1:H}$ are continuous actions and $y^{\ell,a}$ denotes both language tokens as well as discretized action tokens. We use the FAST \citep{pertsch2025fast} tokenizer to convert continuous actions to discrete tokens. We can then jointly sample actions and text from our model, $(a, \hat{\ell}) \sim \pi(\cdot, \cdot | I_{1:V}, q, \ell)$, and train the model with a combination of token prediction (cf. $\mathcal{L}_\text{AR-VLA}$ in \eqref{eq:ar-vla}) and flow matching losses (cf.\ $\mathcal{L}_\text{FLOW-VLA}$ in \eqref{eq:flow-vla}), simultaneously, i.e.\
\vspace{-2mm}
\begin{align}
    \mathcal{L}_{\text{CO-VLA}}(\theta) = \mathbb{E}_{\mathcal{D}, \tau, \omega} \Big[-  \sum_{j=1}^{n-1} M^\ell_j \log p_\theta(\hat{\ell}_{j+1} | x_{1:{j}})
    + \alpha M^\text{act} \left\|\omega - a_{1:H} - f_\theta^a(a^{\tau, \omega}_{1:H})\right\|^2 \Big] \label{eq:cotraining},
\end{align}
where $\alpha$ is a loss multiplier, trading off action prediction via flow-matching with the standard language modeling loss. $M^\ell$ is a language loss mask (indicating locations in the token stream at which the language loss should be applied) and $M^\text{act}$ is an action mask indicator specifying whether or not actions should be predicted for the given example. 
This loss construction allows us to flexibly mix-and-match co-training with data from different modalities. In particular, we combine \textit{VLM} data (which has only images and text annotations) with \textit{action-only} data (where the task is action prediction conditioned on images and text) as well as \textit{combined language and action prediction tasks} (where we take action only data and additionally annotate it with a language description of what the robot should do next)~\citep{Zawalski24-ecot}. As we will see, mixing data of different modalities in this way enhances knowledge transfer in the resulting VLA.
$\hat{\ell}$ contains both text (language) tokens and FAST tokenized action tokens.
Crucially, we set the attention mask $A$ such that no discrete FAST action  token can attend to continuous action tokens and vice-versa. 
We observe in our experiments that this joint training objective lets us combine the best of both worlds: we obtain fast convergence during training from using FAST action tokens to learn good representations, and still obtain fast inference of continuous actions via a few steps of flow-integration. 

\begin{figure}
    \vspace{-2mm}
    \centering
    \includegraphics[width=\linewidth]{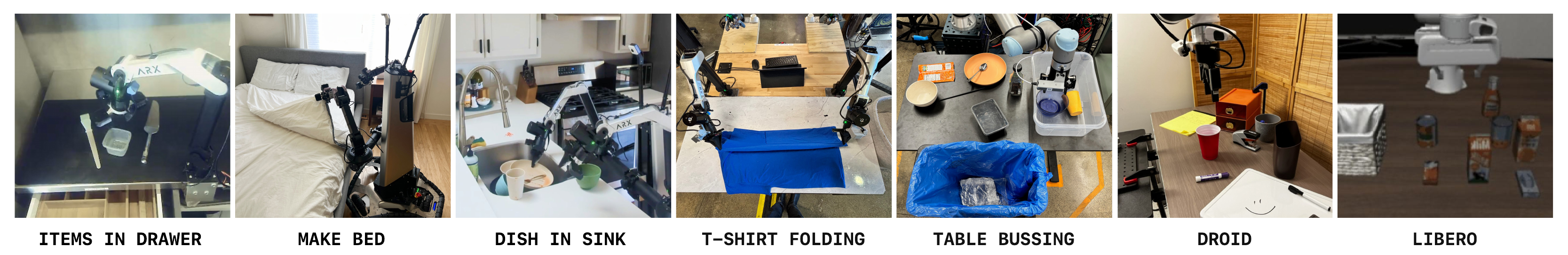}
    \vspace{-6mm}
    \caption{\small Evaluation setups. The left three tasks are evaluated in completely unseen environments.}
    \label{fig:tasks}
    \vspace{-5mm}
\end{figure}

\subsection{Knowledge insulation \& gradient flow}\label{sec:knowledge_insulation}
Gradients from the action expert that is trained with flow matching can unfavorably influence the training dynamics of the image encoder and language model backbone; especially when adding a new, randomly initialized, action expert to a pre-trained backbone.
Therefore, we propose to stop the gradient flow from the action expert to the pre-trained weights in the model.
This is a sensible restriction if and only if the backbone is additionally trained to predict actions directly as part of its language outputs. Since we propose to train the model on discrete actions jointly, we can ensure that the combined activations of the transformer layers contain enough information to infer the actions.
The pre-trained model backbone and action expert only interact via the attention layers. To stop the gradient flow from the action expert into the backbone, we need to modify the attention layers as follows.
For the single head attention case, we can write the attention operation as
$
    P = \mathrm{softmax}\big(Q(X) K(X)^T + A\big) = \begin{pmatrix}
		P_{bb} & 0 \\
		P_{ab} & P_{aa}
	\end{pmatrix}
$    
where X are the inputs to the attention layer, Q, K are the attention query and key projections, respectively,
A is the attention mask as described above, and $\mathrm{softmax}$ is the row-wise softmax. The result are attention probabilities over token features which decompose into probabilities where features from the VLM backbone attend to features from the backbone $P_{bb}$, probabilities for action expert features attending to backbone features $P_{ab}$ and probabilties for action expert features attending other action expert features $P_{aa}$. Given this we can restrict information flow as desired by implementing the softmax computation as
\vspace{-1mm}
\begin{align}
	\begin{pmatrix}
		P_{bb} & 0 \\
		P_{ab} & P_{aa}
	\end{pmatrix}=\mathrm{softmax}\left(\begin{pmatrix}
			Q_b(X_b)K_b(X_b)^T & 0 \\
			Q_a(X_a)\mathrm{sg}\big(K_b(X_b)^T\big) & Q_a(X_a)K_a(X_a)^T
	\end{pmatrix}  + A\right),
\end{align}
where $\mathrm{sg}$ denotes the stop-gradient operator that restricts gradient-flow through this part of the computation.
$X_b$ corresponds to all $x_i$ processed with the backbone weights, $X_a$ to the tokens processed with the action expert weights.
The value embeddings are then computed by
\vspace{-1mm}
\begin{align}
	E = \begin{pmatrix}
		E_b \\
		E_a
	\end{pmatrix} = \begin{pmatrix}
		P_{bb}V_b(X_b)\\
		P_{ab}\mathrm{sg}\big(V_b(X_b)\big) + P_{aa}V_a(X_a)
	\end{pmatrix},
\end{align}
and the final attention is $\mathrm{attn}(X) = P E$.
One additional advantage of this design is that we can simply set $\alpha = 1$ in \eqref{eq:cotraining}, since now the diffusion loss term applies to an independent set of weights.

\section{Experiments}

We evaluate our method on dexterous, long-horizon, manipulation tasks in the real world encompassing multiple different robot embodiments (Figure~\ref{fig:tasks}).
The tasks include cleaning a table (``table bussing''); folding shirts (referred to as ``shirt-folding'') with a bimanual, static robot; putting household items in drawers with a single, static robot arm (``items in drawer''); and multiple tasks involving a bimanual mobile manipulator.
For the latter two, we exclusively evaluate the model in held-out scenes where the model has not seen any data.  
We further show results on the LIBERO simulation benchmark \cite{liu2024libero}, as well as on DROID \cite{khazatsky2024droid} in the real world.
We train models both on single robot embodiments as well as generalist models that are trained on a large mixture of data from many different robots on a large number of tasks, including non-action prediction tasks such as image captioning, bounding box prediction, and robot planning.
We refer to Sec.~\ref{app:dataAndTasks}, \ref{app:trainingDetails} for details on tasks, datasets, and model training.
Our experimental evaluation focuses on the following questions:
\vspace{-0.2cm}
\begin{enumerate}[leftmargin=0.9cm]
    \itemsep0em
    \item \textbf{Performance.} How does our method compare to strong baseline VLAs $\pi_0$ \cite{black2024pi_0}, $\pi_0$-FAST \cite{pertsch2025fast}, HybridVLA \cite{liu2025hybridvla}, OpenVLA-OFT \cite{kim2025fine} in terms of absolute task performance?
    \item \textbf{Knowledge insulation.} What is the effect of stopping the gradient flow?
    \item \textbf{Language following.} A common limitation of many robot policies is that they pay much more attention to images than the language input \cite{kim2025fine}. Which modeling choices influence how well the model pays attention to language inputs, and thus the task at hand?
    \item \textbf{Convergence speed.} How fast does our model train in terms of training steps?
    \item \textbf{Generalization.} Our architecture enables us to train the model not only on robot action data, but also other data sources such as VQA, image captioning, or bounding box prediction. Can we transfer knowledge from these sources into generating actions with the action expert?
    \item \textbf{State representations.} How do different robot state representations influence the model?
\end{enumerate}

We consider the following baselines and ablations \emph{which we re-train on our data mixture}:
\vspace{-0.2cm}
\begin{enumerate}[leftmargin=0.9cm]
    \itemsep0em
    \item $\pi_0$ \cite{black2024pi_0} uses an action expert, continuous actions, and is trained on robot data only.
    \item \texttt{$\pi_0$-FAST} \cite{pertsch2025fast} is an autoregressive VLA with token compression, only trained on robot data.
    \item \texttt{OpenVLA-OFT} \cite{kim2025fine} modifies a standard autoregressive VLA to use parallel decoding with bidirectional attention. We adopt this approach herein, but do not use FiLM and keep text state.
    \item \texttt{Transfusion} \cite{zhou2024transfusion} denoises continuous inputs in the same transformer backbone. The original transfusion work applied their method to image generation. Here we adapt it to robot actions.
    \item \texttt{HybridVLA} \cite{liu2025hybridvla} trains a VLA with transfusion and na\"{i}ve autoregressive tokenization simultaneously. The autoregressive tokens can attend to the diffusion inputs. We slightly modify this architecture to also use an action expert for continuous tokens.
    \item \texttt{joint-training} is the same as our model but without the stop-gradient.
    \item  \texttt{joint-training w/o VLM data}. This ablation removes both the stop-gradient and co-training on VLM data from our proposed method, which can also be considered a variant of HybridVLA \cite{liu2025hybridvla} where we train on both action representations simultaenously, but, compared to HybridVLA, the autoregressive tokens cannot attend to the flow-matching inputs.
    \item \texttt{Naive tokenization} as representation learning objective compared to FAST (see Sec.~\ref{sec:standardVLARecipes}).
\end{enumerate}

\begin{figure}
    \centering
    \captionsetup[subfigure]{aboveskip=0pt, belowskip=0pt}
    \begin{subfigure}[t]{0.48\linewidth}
        \includegraphics[width=\linewidth]{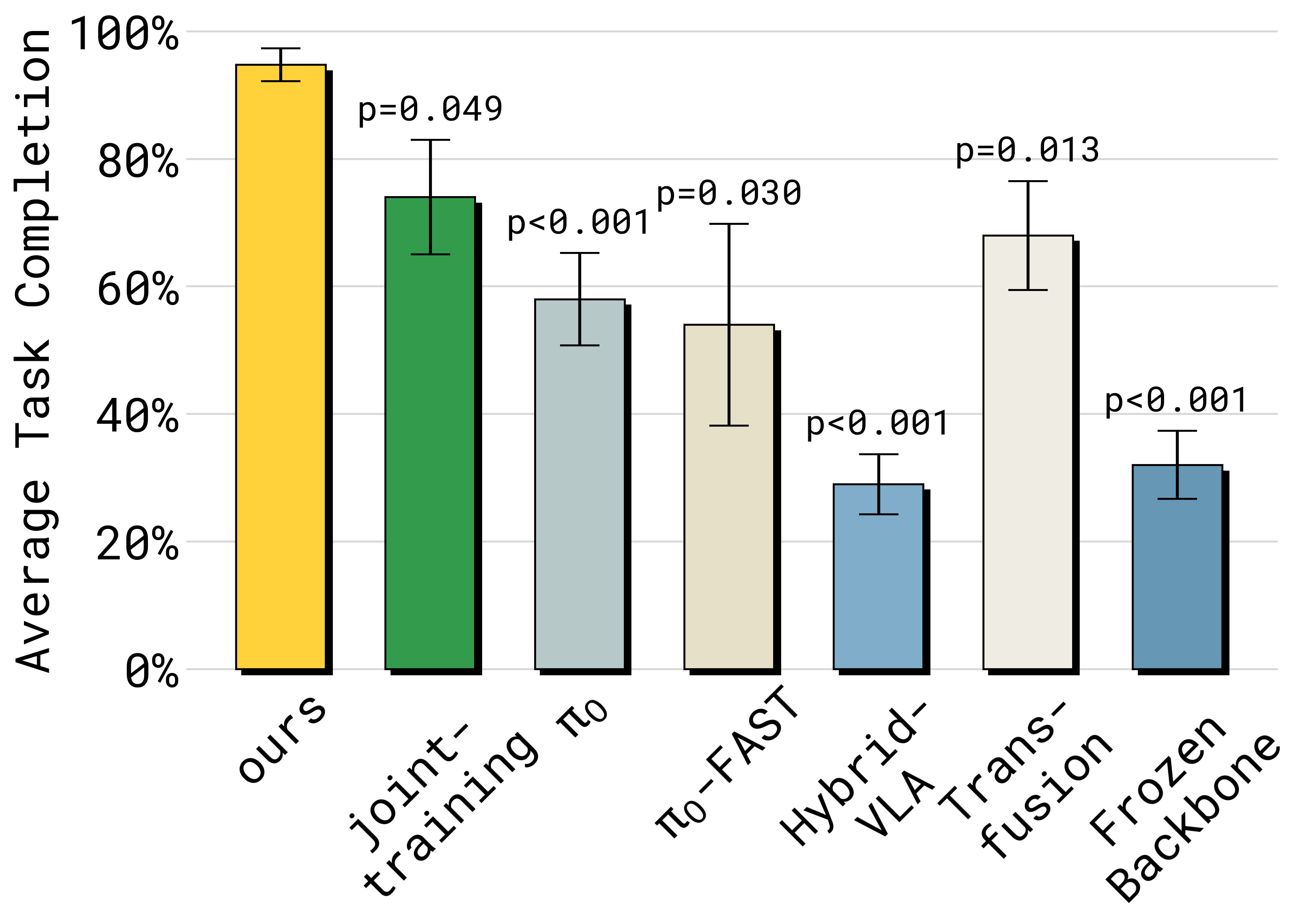}
        \caption{Performance}
        \label{fig:arxSinglePerformance:performance}
    \end{subfigure}
    \hfill
    \begin{subfigure}[t]{0.48\linewidth}
        \includegraphics[width=\linewidth]{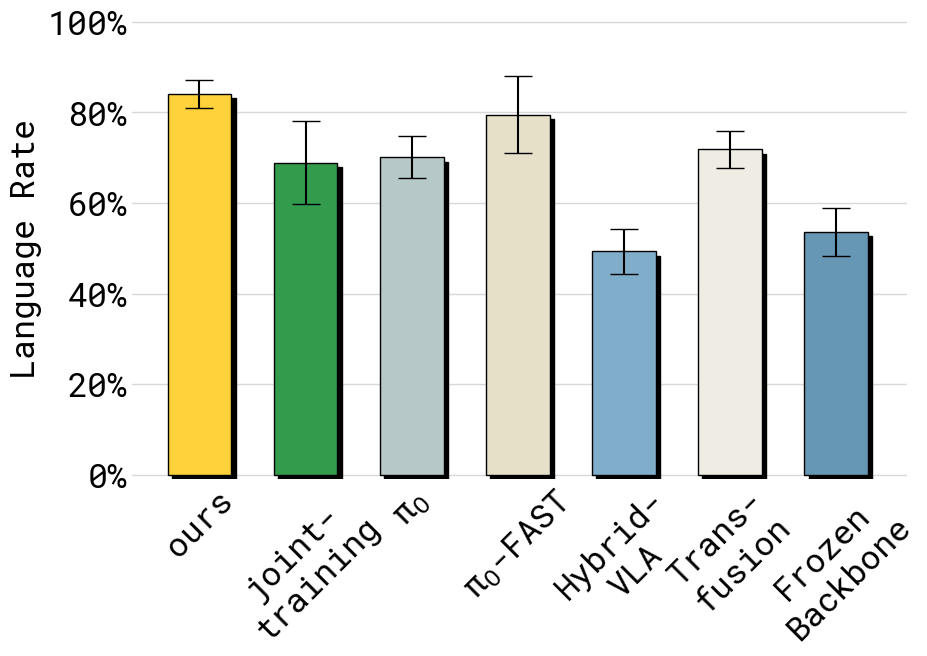}
        \caption{Language following}
        \label{fig:arxSinglePerformance:language}
    \end{subfigure}
    \setlength{\abovecaptionskip}{2pt}
    \caption{\small Comparison to baselines for the  ``items in drawer'' task. Our method outperforms all other baselines both in terms of performance and the ability of the model to follow language instructions. Allowing gradients from the action expert to the backbone (\texttt{joint-training} or $\pi_0$) harms language following. While $\pi_0$\texttt{-FAST} maintains good language following, its performance is worse than our method. Neither \texttt{HybridVLA} nor freezing the backbone is viable for this task.}
    \label{fig:arxSinglePerformance}
\end{figure}

\begin{figure}[ht]
    \centering
    \vspace{-3mm}
    \includegraphics[width=0.8\linewidth]{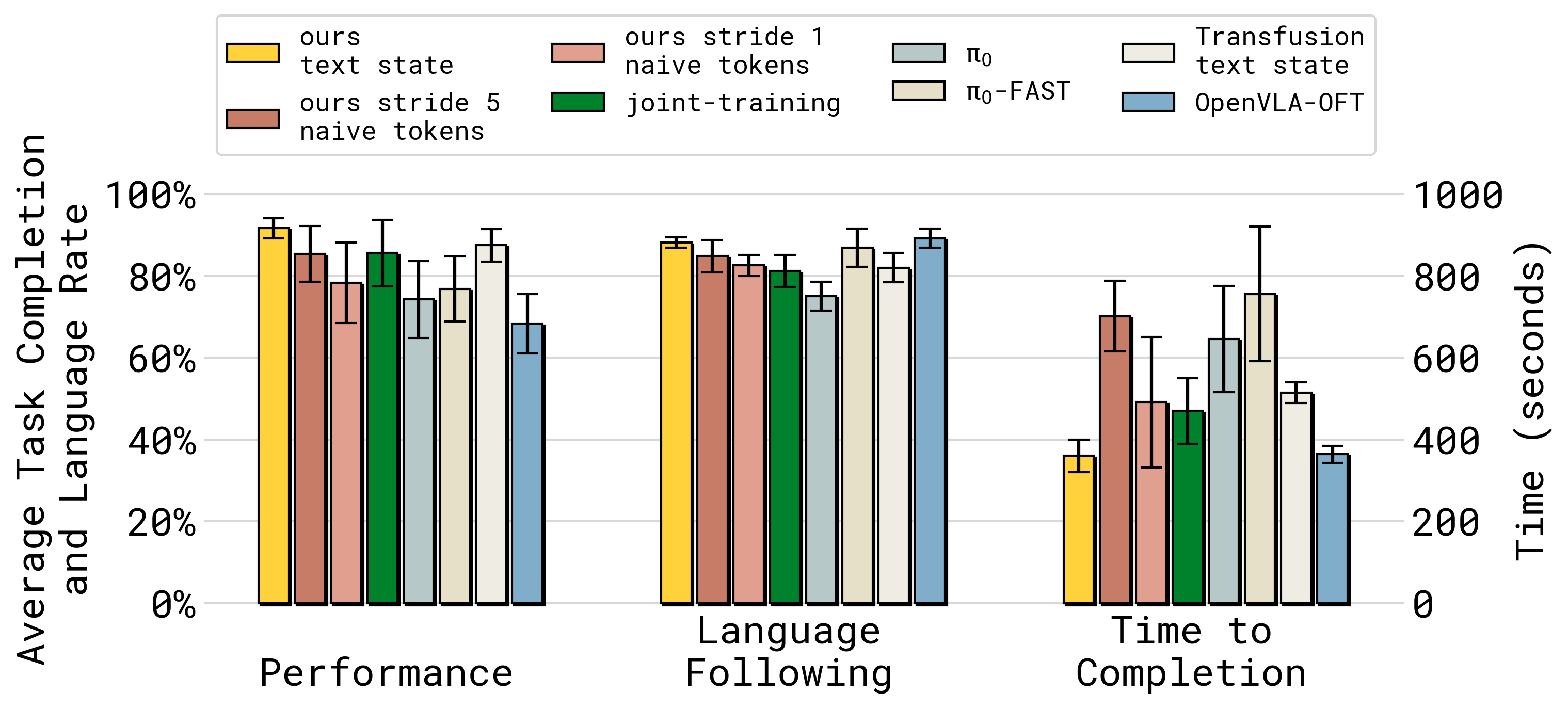}
    \caption{\small Comparison of multiple models/architectures on ``table bussing'' task with specialist models trained on a single robot embodiment. Our model has the highest performance, low inference time, and follows language instructions well. $\pi_0$-FAST also follows language well and has good performance, but requires twice the amount of time (wall clock) to solve the task due to slow inference. $\pi_0$ struggles with following language instructions. OpenVLA-OFT follows language well and has low inference time, but has the lowest overall performance.}
        \label{fig:ur5All}
\end{figure}

\begin{figure}[ht]
    \vspace{-5mm}
    \captionsetup[subfigure]{aboveskip=0pt, belowskip=1pt}
    \begin{subfigure}[t]{0.49\linewidth}
        \centering
        \includegraphics[width=0.65\linewidth]{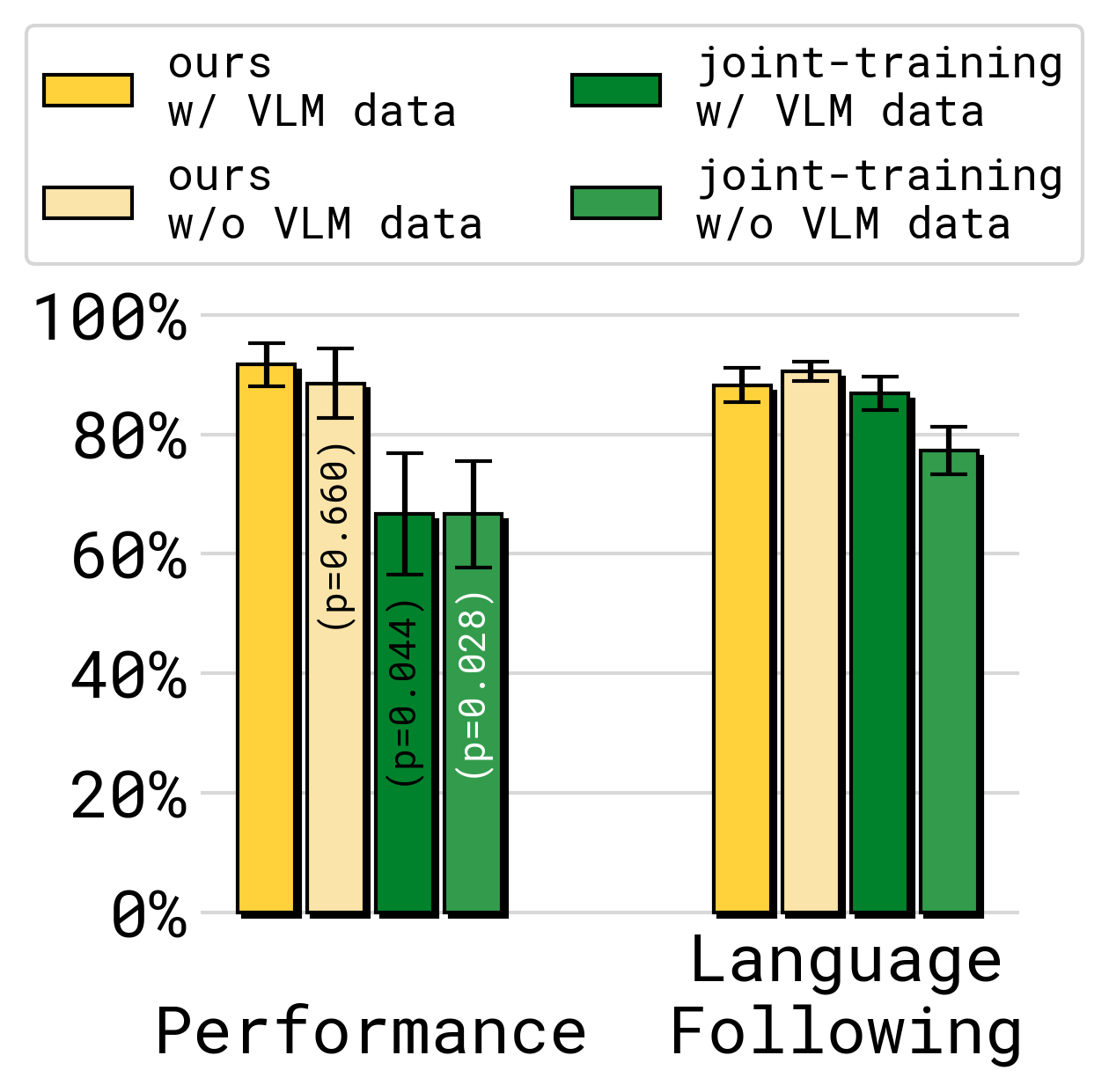}
        \caption{\small Different training strategies}
        \label{fig:ur5Generalist}
    \end{subfigure}
    \hfill
    \begin{subfigure}[t]{0.5\linewidth}
        \centering
        \includegraphics[width=0.8\linewidth]{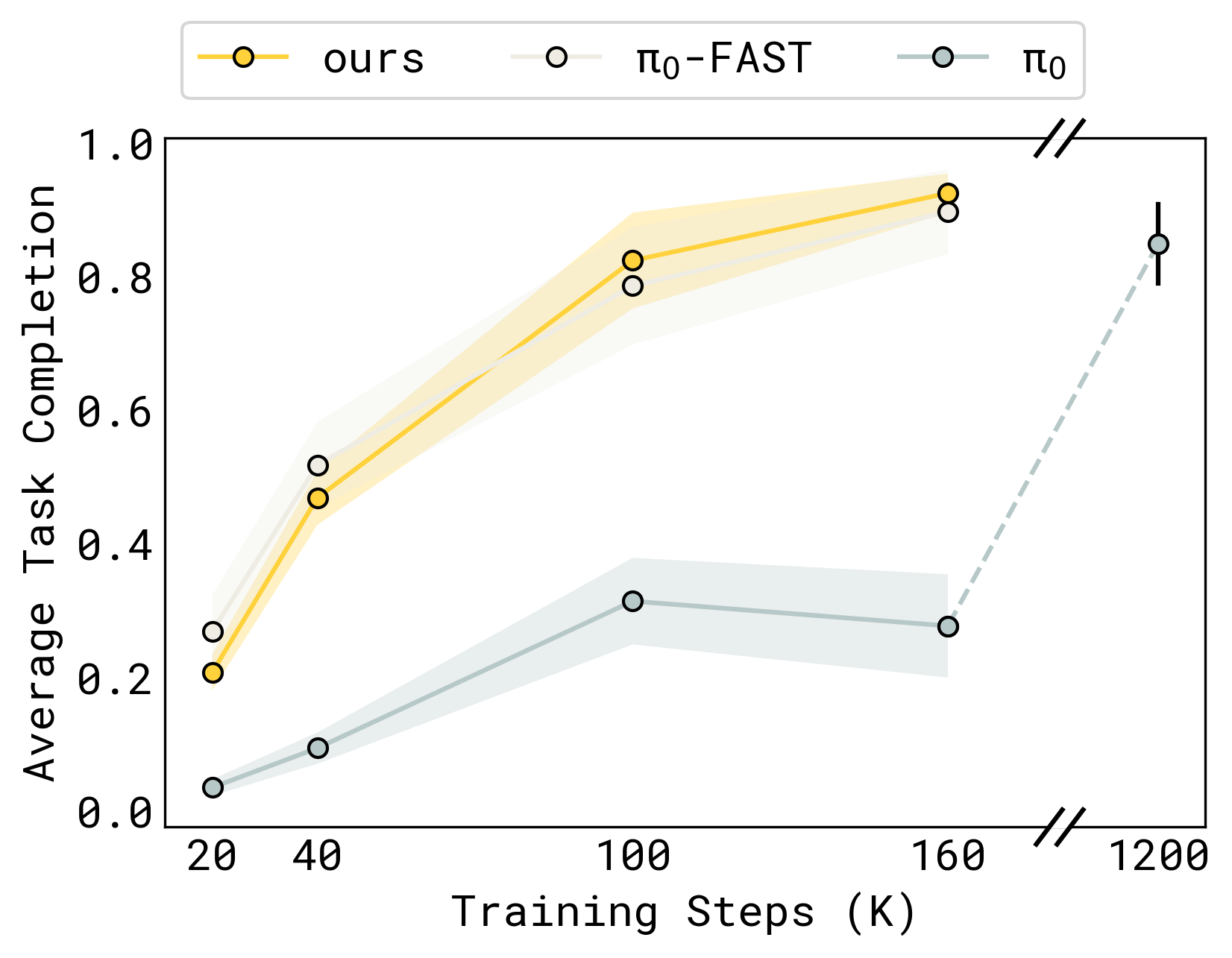}
        \caption{\small Performance over number of training steps}
        \label{fig:UR5TrainingSpeed}
    \end{subfigure}
    \setlength{\abovecaptionskip}{2pt}
    \caption{Results on ``table bussing'' task with generalist model trained on many embodiments. Our model follows language well, and trains as quickly as $\pi_0$-FAST. In comparison, $\pi_0$ trains significantly slower, requiring 7.5 times as many training steps to reach a similar performance.}
    \label{fig:ur5}
\end{figure}

\setlength{\textfloatsep}{6pt}
\begin{figure}
    \centering
    \begin{minipage}{0.58\linewidth}
        \includegraphics[width=1.0\linewidth]{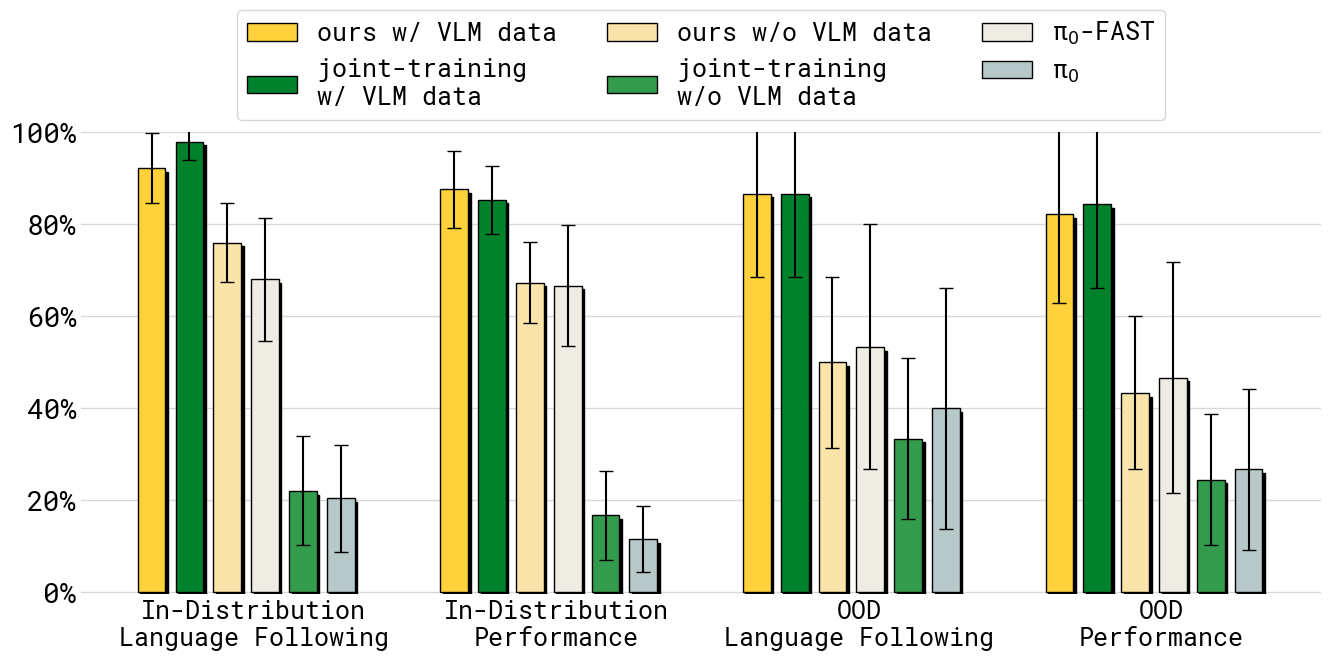}%
        \setlength{\abovecaptionskip}{1pt}
        \caption{\small Generalization to novel objects (mobile manipulator).}
        \label{fig:MMLanguageFollowing}
    \end{minipage}
    \hspace{-0.2cm}
    \begin{minipage}{0.42\linewidth}
        \includegraphics[width=1.0\linewidth]{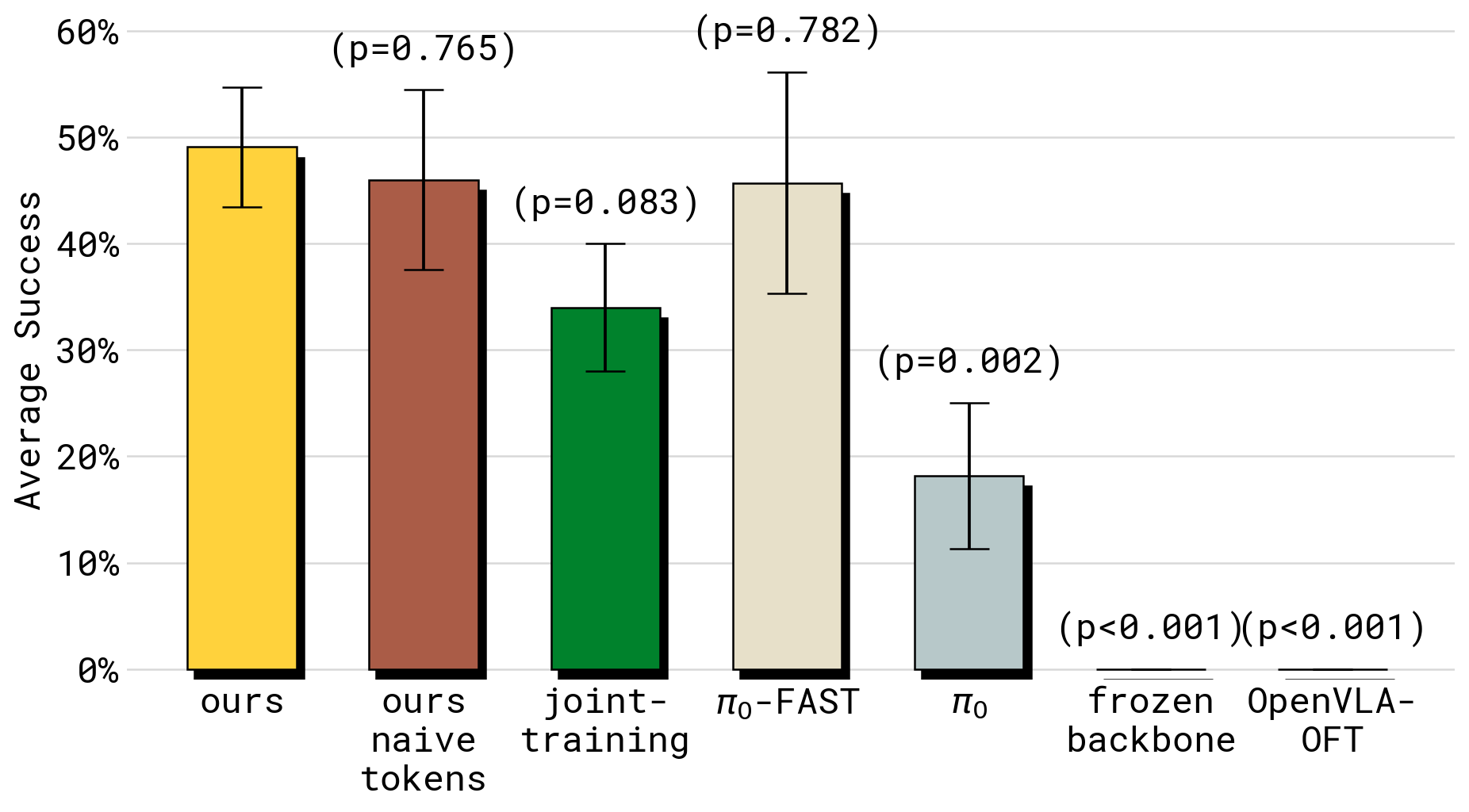}%
        \setlength{\abovecaptionskip}{1pt}
        \caption{\small Performance on ``shirt folding''.}
        \label{fig:shirtFolding}
    \end{minipage}
\end{figure}

\begin{figure}
    \vspace{-3mm}
    \begin{minipage}{0.4\textwidth}
        \centering
        \includegraphics[width=0.94\textwidth]{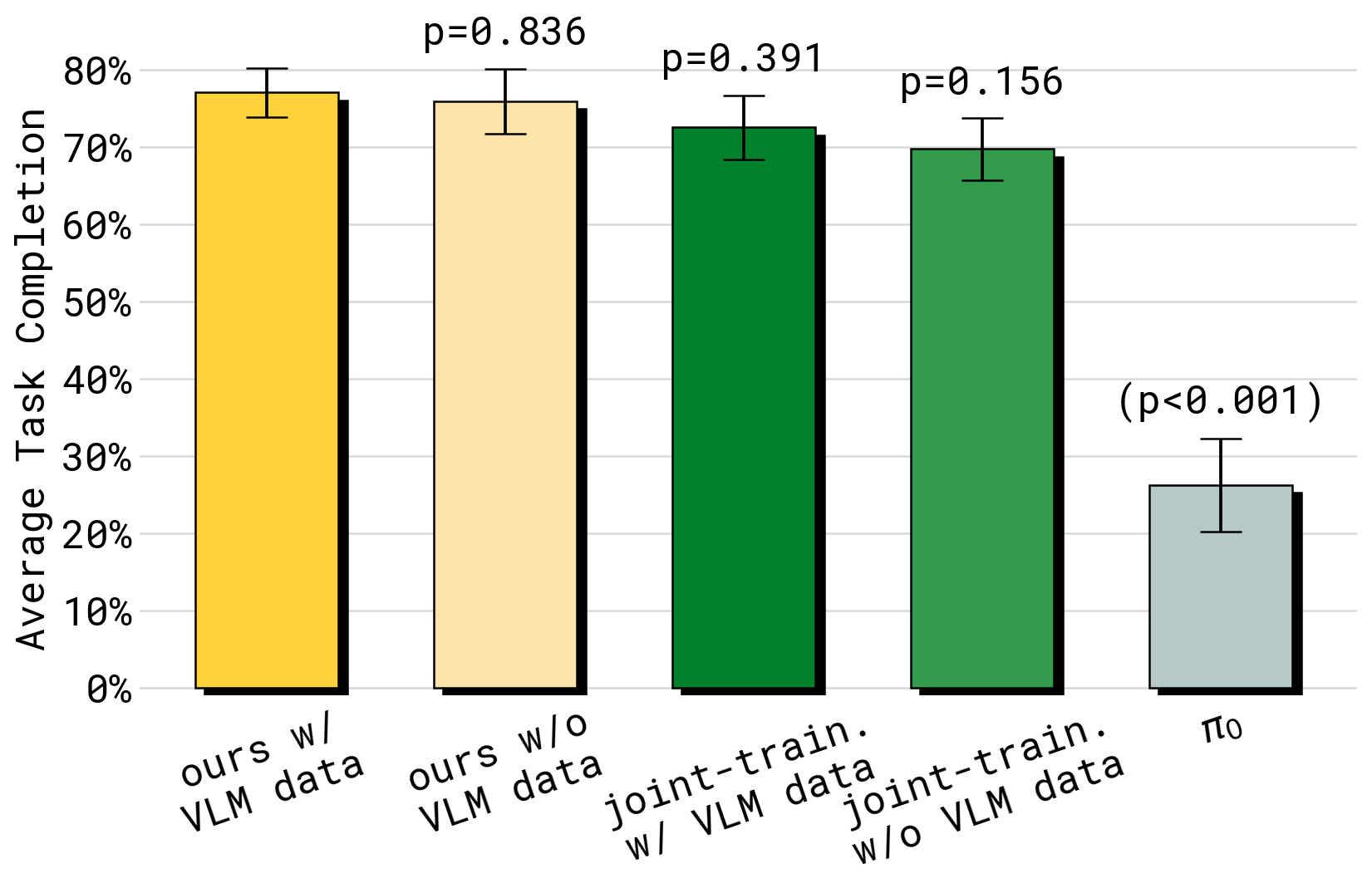}
        \vspace{-3mm}
        \caption{\small Average performance on 4 mobile manipulation tasks in unseen environments.}
        \label{fig:mobilePerformance}
    \end{minipage}
    \hfill
    \begin{minipage}{0.57\textwidth}
    \centering
        \resizebox{0.8\textwidth}{!}{
        \begin{tabular}{l@{}c@{\hspace{0.15cm}}c@{\hspace{0.15cm}}c@{\hspace{0.15cm}}c@{\hspace{0.15cm}}c}
        \toprule
         & Spatial & Object & Goal & 10 (Long) & 90 \\
        \midrule
        Baku \cite{haldar2024baku} & -- & -- & -- & 86.0 & 90.0 \\
        MoDE \cite{reuss2024efficient} & -- & -- & -- & 94.0 & 95.0 \\
        OpenVLA-OFT \cite{kim2025fine} & 97.6 & 98.4 & \textbf{97.9} & \textbf{94.5} & -- \\
        $\pi_0$ \cite{black2024pi_0} & 96.8 & \textbf{98.8} & 95.8 & 85.2 & -- \\
        $\pi_0$-FAST \cite{pertsch2025fast} & 96.4 & 96.8 & 88.6 & 60.2 & -- \\
        Ours (from scratch) & 96.6 & 97.2 & 94.6 & 84.8 & 92.7 \\
        Ours (from generalist model) & \textbf{98.0} & 97.8 & 95.6 & 85.8 & \textbf{96.0} \\
        \bottomrule
        \end{tabular}
        }
        \captionof{table}{\small Success rates (\%) on the LIBERO \cite{liu2024libero} benchmark. Our method achieves a state-of-the-art in LIBERO-90 and LIBERO-Spatial, but is worse on LIBERO-10.}
        \label{tab:libero}
    \end{minipage}
    \vspace{-1mm}
\end{figure}

\textbf{Task performance \& comparison to baselines.}
Our method consistently achieves the highest performance in the real world evaluations.
For the ``items in drawer'' task, which requires both accurate language following (to pick the right object) and precise manipulation (to open a kitchen drawer), all baselines perform significantly worse than our proposed approach (Fig.~\ref{fig:arxSinglePerformance:performance}) with a common failure mode of being unable to open the drawer. Note that this task is evaluated in a held-out environment. In particular, the \texttt{joint-training} baseline (no stop gradient) has problems following language, similar issues occur with $\pi_0$. \texttt{$\pi_0$-FAST} moves slowly and fails to open the drawer with precision in many cases.
\texttt{HybridVLA} \cite{liu2025hybridvla} is the baseline that is methodologically most similar to our approach, since it also jointly trains on both discretized and continuous actions, but allows autoregressive tokens to attend to continuous actions. This seems to hurt performance on this task significantly. Setting the attention mask as we propose leads to much better performance.
A detailed ablation of modeling choices made for our method as well as other baselines on the ``table bussing'' task is depicted in Fig.~\ref{fig:ur5All}. As before, our method performs best, here \texttt{joint-training} also performs well. \texttt{$\pi_0$-FAST} is slow, requiring \emph{twice} the amount of time to solve the task. \texttt{Transfusion} performs well but is slower than our method.
Using parallel decoding (\texttt{OpenVLA-OFT}) also generally performs worse.
Freezing the backbone is \emph{not} a viable option for knowledge insulation, since the representations in the pre-trained model are not sufficient for robotics, leading to low performance cf. Fig.~\ref{fig:arxSinglePerformance:performance} and shirt-folding,  Fig.~\ref{fig:shirtFolding}, where $\pi_0$ also struggles due to being trained on our large single embodiment data-mix. Again in this setting freezing the backbone or parallel decoding are not good strategies.
We also evaluate our generalist on the open source benchmark DROID \cite{khazatsky2024droid} for the same set of tabletop manipulation tasks as in \citep{pertsch2025fast}.
Our method received a score of $0.55 \pm 0.09$, $\pi_0$ received $0.49 \pm 0.09$, and $\pi_0$-FAST achieved $0.45 \pm 0.09$.
Finally, our approach achieves a new state-of-the-art in LIBERO-90 and LIBERO-Spatial \cite{liu2024libero} as shown in Tab.~\ref{tab:libero}.
This model has been finetuned on LIBERO from the generalist stop-gradient + VLM data co-trained model since
the generalist model did not have LIBERO data in its original training mixture.

\textbf{Generalist VLA evaluation.}
While the previous results considered VLAs trained with data from the target embodiment only (though with more tasks than we evaluated for) we next shift to assessing how well our recipe works when training jointly on all data we have available for training. Fig.~\ref{fig:ur5Generalist} shows that for the ``table bussing" task our recipe achieves comparable performance to the embodiment specific results from above. In comparison \texttt{joint-training} degrades in task completion. We also show that removing VLM data (e.g. \texttt{ours w/o VLM data}) leads to slightly worse task completion percentage. Interestingly when looking at the rate at which the policy follows the human provided language commands for cleaning the table, removing VLM data has the biggest effect on \texttt{joint-training}. We hypothesize that this data is especially needed to avoid catastrophic interference with pre-trained representations in this case. Finally, we evaluate our method on four mobile manipulation tasks (e.g.. placing dishes in a sink, see Sec.~\ref{sec:tasksMobileManipulator} for details). Results are shown in Fig. 
\ref{fig:mobilePerformance} where similar trends emerge and our method trained with VLM data performs best. Notably $\pi_0$ performs worse when evaluated after the same number of training steps; we elucidate why in Fig. \ref{fig:UR5TrainingSpeed} where we can see that training with flow-matching loss only ($\pi_0$) requires many more steps to converge whereas training with our method is as quick as training with FAST.

\textbf{Language following.}
In any scene, a robot can typically execute many sensible actions, for example grasping different objects.
Here we evaluate whether different models, when given a specific task (provided via natural language instructions), produces actions that attempt to achieve this task.
This is particularly important for tasks which have a long-horizon goals such as cleaning a kitchen counter where models could easily overfit on provided data to solving the long-horizon task by focussing on the image inputs alone.
We hypothesize that if a VLA maintains more of its VLM pre-training knowledge, it should be more likely to pay attention to the actual language input.
As one can see in Fig.~\ref{fig:arxSinglePerformance:language}, stopping the gradient flow from the action expert is an effective way of improving language following compared to $\pi_0$ and \texttt{joint-training} without stop-gradient and without VLM data co-training.
As mentioned above already, if the model is co-trained with VLM data, as shown in Fig.~\ref{fig:ur5Generalist}, Fig.~\ref{fig:ur5All}, and Fig.~\ref{fig:MMLanguageFollowing}, then \texttt{joint-training} without stop-gradient can also achieve good language following.
Further, \texttt{transfusion} (Fig.~\ref{fig:ur5All}) follows language better than $\pi_0$ with the action expert, which can be explained by the fact that it reuses the backbone weights for continuous action generation and the only newly initialized parameters are the action projections.
These results strongly support the hypothesis that gradients from randomly initialized robotics specific adapters unfavorably interact with the pre-trained VLM weights.
Our proposed knowledge insulation techniques of stoping gradient flow, and/or co-training with VLM data are able to achieve better language following.

\textbf{Transfer from VLM data into robotics.}
One of the main motivations for VLAs is transferring knowledge from non-robot data sources to robot policies.
We perform an experiment with a mobile manipulation robot, where we investigate semantic generalization to new objects.
The robot is tasked with moving objects from a kitchen counter into an (already open) drawer.
The objects are not seen during training.
As one can see in Fig.~\ref{fig:MMLanguageFollowing} under ``OOD Follow Rate'', co-training on VLM data is particularly important for this generalization. 

\textbf{Investigation of other modeling choices.}
Our main recipe uses FAST \cite{pertsch2025fast} to represent discrete actions as a representation learning objective for the model backbone.
One motivation for FAST is that it provides a better learning signal compared to naive tokenization, and can help with faster inference (due to fewer tokens).
Since here we use the discrete action tokens only during training time, one may wonder whether simpler, naive tokenization is sufficient for learning good representations.
To investigate, we exchange FAST with naive tokenization during training, but keep all other choices the same.
The resulting model is still better than training with continuous actions alone, but worse than using FAST for representation learning (Fig.~\ref{fig:ur5All}) (cf. \texttt{naive tokens} in the figure).
Sub-sampling tokens (via stride of 5), in this case is better than dense naive tokenization.
In the Appendix (Fig.~\ref{fig:stateRepresentations}), we compare different choices for the robot's proprioceptive state, cf.\ Sec.~\ref{sec:stateRepresentations}, on the table bussing task and find that our method works both with text and continuous state.

\section{Discussion \& Limitations}

We analyze the performance, generalization, and language following capabilities of continuous-action VLAs that fine-tune VLMs to output continuous actions, show that such models suffer from a significant loss of pre-trained knowledge, and propose a method that can greatly mitigate this degradation by shielding the pre-trained VLM backbone during VLA training. The core idea in our approach is to use discretized actions to provide a learning signal to fine-tune VLM representations, while simultaneously training a continuous (flow matching) action expert \emph{without} propagating its gradient into the VLM. The VLM backbone is thus not damaged by backpropagation from the action expert but still receives a learning signal (from discrete actions) that adapts its representations to the robot control task. Experiments across numerous real-world and simulated tasks provide strong evidence for our hypothesis about the degradation of the VLM backbone with na\"{i}ve training, and a clear indication that our approach mitigates this challenge.

Our method provides an effective recipe for training continuous-action VLAs, but does have limitations. Training with both continuous and discrete outputs increases computational cost by about 20\% during training. However, due to the increased convergence speed, this cost is offset such that our model still trains much faster (in wall-clock time) relative to purely diffusion based VLAs such as $\pi_0$. Additionally, while our method improves language following it is still far from perfect, likely because correlations in the training data still cause the model to sometimes ignore language instructions.

\paragraph{Acknowledgments}
We thank Chelsea Finn for initial experiments regarding language following;
Ury Zhilinsky, Karan Dhabalia, Haohuan Wang, Dibya Gosh, Kyle Stachowicz, Kevin Black for training infrastructure;
Noah Brown, Szymon Jakubczak, Adnan Esmail, Tim Jones, Mohith Mothukuri, James Darpinian, James Tanner for help with robot infrastructure;
and Anna Walling, Chelsea Finn, Karol Hausman for help with robot, data and eval operations.
We are grateful to the whole team of robot operators at Physical Intelligence for their enormous contributions to running data collection and policy evaluations. Finally, we thank Claudio Guglieri, Alex Krasikov, Spike Brehm, Lachy Groom and Karol Hausman
for their help with visualizations for the website.

%%%%%%%%%%%%%%%%%%%%%%%%%%%%%%%%%%%%%%%%%%%%%%%%%%%%%%%%%%%%

\bibliographystyle{plainnat}
\bibliography{references}

\begin{thebibliography}{59}
\providecommand{\natexlab}[1]{#1}
\providecommand{\url}[1]{\texttt{#1}}
\expandafter\ifx\csname urlstyle\endcsname\relax
  \providecommand{\doi}[1]{doi: #1}\else
  \providecommand{\doi}{doi: \begingroup \urlstyle{rm}\Url}\fi

\bibitem[AgiBot-World-Contributors et~al.(2025)AgiBot-World-Contributors, Bu,
  Cai, Chen, Cui, Ding, Feng, Gao, He, Hu, Huang, Jiang, Jiang, Jing, Li, Li,
  Liu, Liu, Lu, Luo, Luo, Mu, Niu, Pan, Pang, Qiao, Ren, Ruan, Shan, Shen, Shi,
  Shi, Shi, Sima, Song, Wang, Wang, Wei, Xie, Xu, Yan, Yang, Yang, Yang, Yao,
  Zeng, Zhang, Zhang, Zhao, Zhao, Zhao, and Zhu]{contributors2025agibotworld}
AgiBot-World-Contributors, Qingwen Bu, Jisong Cai, Li~Chen, Xiuqi Cui, Yan
  Ding, Siyuan Feng, Shenyuan Gao, Xindong He, Xuan Hu, Xu~Huang, Shu Jiang,
  Yuxin Jiang, Cheng Jing, Hongyang Li, Jialu Li, Chiming Liu, Yi~Liu, Yuxiang
  Lu, Jianlan Luo, Ping Luo, Yao Mu, Yuehan Niu, Yixuan Pan, Jiangmiao Pang,
  Yu~Qiao, Guanghui Ren, Cheng Ruan, Jiaqi Shan, Yongjian Shen, Chengshi Shi,
  Mingkang Shi, Modi Shi, Chonghao Sima, Jianheng Song, Huijie Wang, Wenhao
  Wang, Dafeng Wei, Chengen Xie, Guo Xu, Junchi Yan, Cunbiao Yang, Lei Yang,
  Shukai Yang, Maoqing Yao, Jia Zeng, Chi Zhang, Qinglin Zhang, Bin Zhao,
  Chengyue Zhao, Jiaqi Zhao, and Jianchao Zhu.
\newblock Agibot world colosseo: A large-scale manipulation platform for
  scalable and intelligent embodied systems.
\newblock \emph{arXiv preprint arXiv:2503.06669}, 2025.

\bibitem[Alayrac et~al.(2022)Alayrac, Donahue, Luc, Miech, Barr, Hasson, Lenc,
  Mensch, Millican, Reynolds, et~al.]{alayrac2022flamingo}
Jean-Baptiste Alayrac, Jeff Donahue, Pauline Luc, Antoine Miech, Iain Barr,
  Yana Hasson, Karel Lenc, Arthur Mensch, Katherine Millican, Malcolm Reynolds,
  et~al.
\newblock Flamingo: a visual language model for few-shot learning.
\newblock \emph{Advances in neural information processing systems},
  35:\penalty0 23716--23736, 2022.

\bibitem[Belkhale and Sadigh(2024)]{belkhale2024minivla}
Suneel Belkhale and Dorsa Sadigh.
\newblock Minivla: A better vla with a smaller footprint, 2024.
\newblock URL \url{https://github.com/Stanford-ILIAD/openvla-mini}.

\bibitem[Beyer et~al.(2024)Beyer, Steiner, Pinto, Kolesnikov, Wang, Salz,
  Neumann, Alabdulmohsin, Tschannen, Bugliarello, et~al.]{beyer2024paligemma}
Lucas Beyer, Andreas Steiner, Andr{\'e}~Susano Pinto, Alexander Kolesnikov,
  Xiao Wang, Daniel Salz, Maxim Neumann, Ibrahim Alabdulmohsin, Michael
  Tschannen, Emanuele Bugliarello, et~al.
\newblock Paligemma: A versatile 3b vlm for transfer.
\newblock \emph{arXiv preprint arXiv:2407.07726}, 2024.

\bibitem[Bharadhwaj et~al.(2024)Bharadhwaj, Vakil, Sharma, Gupta, Tulsiani, and
  Kumar]{bharadhwaj2023roboagent}
Homanga Bharadhwaj, Jay Vakil, Mohit Sharma, Abhinav Gupta, Shubham Tulsiani,
  and Vikash Kumar.
\newblock Roboagent: Generalization and efficiency in robot manipulation via
  semantic augmentations and action chunking.
\newblock In \emph{2024 IEEE International Conference on Robotics and
  Automation (ICRA)}, pages 4788--4795. IEEE, 2024.

\bibitem[Bjorck et~al.(2025)Bjorck, Casta{\~n}eda, Cherniadev, Da, Ding, Fan,
  Fang, Fox, Hu, Huang, et~al.]{bjorck2025gr00t}
Johan Bjorck, Fernando Casta{\~n}eda, Nikita Cherniadev, Xingye Da, Runyu Ding,
  Linxi Fan, Yu~Fang, Dieter Fox, Fengyuan Hu, Spencer Huang, et~al.
\newblock Gr00t n1: An open foundation model for generalist humanoid robots.
\newblock \emph{arXiv preprint arXiv:2503.14734}, 2025.

\bibitem[Black et~al.(2024)Black, Brown, Driess, Esmail, Equi, Finn, Fusai,
  Groom, Hausman, Ichter, Jakubczak, Jones, Ke, Levine, Li-Bell, Mothukuri,
  Nair, Pertsch, Shi, Tanner, Vuong, Walling, Wang, and
  Zhilinsky]{black2024pi_0}
Kevin Black, Noah Brown, Danny Driess, Adnan Esmail, Michael Equi, Chelsea
  Finn, Niccolo Fusai, Lachy Groom, Karol Hausman, Brian Ichter, Szymon
  Jakubczak, Tim Jones, Liyiming Ke, Sergey Levine, Adrian Li-Bell, Mohith
  Mothukuri, Suraj Nair, Karl Pertsch, Lucy~Xiaoyang Shi, James Tanner, Quan
  Vuong, Anna Walling, Haohuan Wang, and Ury Zhilinsky.
\newblock $\pi_0$: A vision-language-action flow model for general robot
  control.
\newblock \emph{arXiv preprint arXiv:2410.24164}, 2024.

\bibitem[Bu et~al.(2025)Bu, Cai, Chen, Cui, Ding, Feng, Gao, He, Huang, Jiang,
  et~al.]{bu2025agibot}
Qingwen Bu, Jisong Cai, Li~Chen, Xiuqi Cui, Yan Ding, Siyuan Feng, Shenyuan
  Gao, Xindong He, Xu~Huang, Shu Jiang, et~al.
\newblock Agibot world colosseo: A large-scale manipulation platform for
  scalable and intelligent embodied systems.
\newblock \emph{arXiv preprint arXiv:2503.06669}, 2025.

\bibitem[Chen et~al.(2022)Chen, Guo, Yi, Li, and Elhoseiny]{chen2022visualgpt}
Jun Chen, Han Guo, Kai Yi, Boyang Li, and Mohamed Elhoseiny.
\newblock Visualgpt: Data-efficient adaptation of pretrained language models
  for image captioning.
\newblock In \emph{Proceedings of the IEEE/CVF conference on computer vision
  and pattern recognition}, pages 18030--18040, 2022.

\bibitem[Chen et~al.(2015)Chen, Fang, Lin, Vedantam, Gupta, Doll{\'a}r, and
  Zitnick]{chen2015microsoft}
Xinlei Chen, Hao Fang, Tsung-Yi Lin, Ramakrishna Vedantam, Saurabh Gupta, Piotr
  Doll{\'a}r, and C~Lawrence Zitnick.
\newblock Microsoft coco captions: Data collection and evaluation server.
\newblock \emph{arXiv preprint arXiv:1504.00325}, 2015.

\bibitem[Chi et~al.(2023)Chi, Xu, Feng, Cousineau, Du, Burchfiel, Tedrake, and
  Song]{chi2023diffusion}
Cheng Chi, Zhenjia Xu, Siyuan Feng, Eric Cousineau, Yilun Du, Benjamin
  Burchfiel, Russ Tedrake, and Shuran Song.
\newblock Diffusion policy: Visuomotor policy learning via action diffusion.
\newblock \emph{The International Journal of Robotics Research}, page
  02783649241273668, 2023.

\bibitem[Dasari et~al.(2019)Dasari, Ebert, Tian, Nair, Bucher, Schmeckpeper,
  Singh, Levine, and Finn]{dasari2019robonet}
Sudeep Dasari, Frederik Ebert, Stephen Tian, Suraj Nair, Bernadette Bucher,
  Karl Schmeckpeper, Siddharth Singh, Sergey Levine, and Chelsea Finn.
\newblock Robonet: Large-scale multi-robot learning.
\newblock \emph{CoRL}, 2019.

\bibitem[Deitke et~al.(2024)Deitke, Clark, Lee, Tripathi, Yang, Park, Salehi,
  Muennighoff, Lo, Soldaini, et~al.]{deitke2024molmo}
Matt Deitke, Christopher Clark, Sangho Lee, Rohun Tripathi, Yue Yang, Jae~Sung
  Park, Mohammadreza Salehi, Niklas Muennighoff, Kyle Lo, Luca Soldaini, et~al.
\newblock Molmo and pixmo: Open weights and open data for state-of-the-art
  multimodal models.
\newblock \emph{arXiv preprint arXiv:2409.17146}, 2024.

\bibitem[Driess et~al.(2023)Driess, Xia, Sajjadi, Lynch, Chowdhery, Ichter,
  Wahid, Tompson, Vuong, Yu, et~al.]{driess2023palm}
Danny Driess, Fei Xia, Mehdi~SM Sajjadi, Corey Lynch, Aakanksha Chowdhery,
  Brian Ichter, Ayzaan Wahid, Jonathan Tompson, Quan Vuong, Tianhe Yu, et~al.
\newblock Palm-e: An embodied multimodal language model.
\newblock \emph{arXiv preprint arXiv:2303.03378}, 2023.

\bibitem[Ebert et~al.(2021)Ebert, Yang, Schmeckpeper, Bucher, Georgakis,
  Daniilidis, Finn, and Levine]{ebert2021bridge}
Frederik Ebert, Yanlai Yang, Karl Schmeckpeper, Bernadette Bucher, Georgios
  Georgakis, Kostas Daniilidis, Chelsea Finn, and Sergey Levine.
\newblock Bridge data: Boosting generalization of robotic skills with
  cross-domain datasets.
\newblock \emph{arXiv preprint arXiv:2109.13396}, 2021.

\bibitem[Esser et~al.(2024)Esser, Kulal, Blattmann, Entezari, M{\"u}ller,
  Saini, Levi, Lorenz, Sauer, Boesel, et~al.]{esser2024scaling}
Patrick Esser, Sumith Kulal, Andreas Blattmann, Rahim Entezari, Jonas
  M{\"u}ller, Harry Saini, Yam Levi, Dominik Lorenz, Axel Sauer, Frederic
  Boesel, et~al.
\newblock Scaling rectified flow transformers for high-resolution image
  synthesis.
\newblock In \emph{Forty-first International Conference on Machine Learning},
  2024.

\bibitem[Fang et~al.(2024)Fang, Fang, Tang, Liu, Wang, Wang, Zhu, and
  Lu]{fang2024rh20t}
Hao-Shu Fang, Hongjie Fang, Zhenyu Tang, Jirong Liu, Chenxi Wang, Junbo Wang,
  Haoyi Zhu, and Cewu Lu.
\newblock Rh20t: A comprehensive robotic dataset for learning diverse skills in
  one-shot.
\newblock In \emph{2024 IEEE International Conference on Robotics and
  Automation (ICRA)}, pages 653--660. IEEE, 2024.

\bibitem[Gage(1994)]{gage1994new}
Philip Gage.
\newblock A new algorithm for data compression.
\newblock \emph{The C Users Journal}, 12\penalty0 (2):\penalty0 23--38, 1994.

\bibitem[Goyal et~al.(2017)Goyal, Khot, Summers-Stay, Batra, and
  Parikh]{goyal2017making}
Yash Goyal, Tejas Khot, Douglas Summers-Stay, Dhruv Batra, and Devi Parikh.
\newblock Making the {V} in {VQA} matter: Elevating the role of image
  understanding in visual question answering.
\newblock In \emph{Computer Vision and Pattern Recognition (CVPR)}, 2017.

\bibitem[Haldar et~al.(2024)Haldar, Peng, and Pinto]{haldar2024baku}
Siddhant Haldar, Zhuoran Peng, and Lerrel Pinto.
\newblock Baku: An efficient transformer for multi-task policy learning.
\newblock \emph{arXiv preprint arXiv:2406.07539}, 2024.

\bibitem[Huang et~al.(2025)Huang, Liu, Fu, Wu, Mukadam, Malik, Goldberg, and
  Abbeel]{huang2025otter}
Huang Huang, Fangchen Liu, Letian Fu, Tingfan Wu, Mustafa Mukadam, Jitendra
  Malik, Ken Goldberg, and Pieter Abbeel.
\newblock Otter: A vision-language-action model with text-aware visual feature
  extraction.
\newblock \emph{arXiv preprint arXiv:2503.03734}, 2025.

\bibitem[Intelligence et~al.(2025)Intelligence, Black, Brown, Darpinian,
  Dhabalia, Driess, Esmail, Equi, Finn, Fusai, et~al.]{pi2025pi05}
Physical Intelligence, Kevin Black, Noah Brown, James Darpinian, Karan
  Dhabalia, Danny Driess, Adnan Esmail, Michael Equi, Chelsea Finn, Niccolo
  Fusai, et~al.
\newblock $\pi_{0.5}$: a vision-language-action model with open-world
  generalization.
\newblock \emph{arXiv preprint arXiv:2504.16054}, 2025.

\bibitem[Khazatsky et~al.(2024)Khazatsky, Pertsch, Nair, Balakrishna, Dasari,
  Karamcheti, Nasiriany, Srirama, Chen, Ellis, Fagan, Hejna, Itkina, Lepert,
  Ma, Miller, Wu, Belkhale, Dass, Ha, Jain, Lee, Lee, Memmel, Park,
  Radosavovic, Wang, Zhan, Black, Chi, Hatch, Lin, Lu, Mercat, Rehman, Sanketi,
  Sharma, Simpson, Vuong, Walke, Wulfe, Xiao, Yang, Yavary, Zhao, Agia, Baijal,
  Castro, Chen, Chen, Chung, Drake, Foster, Gao, Herrera, Heo, Hsu, Hu,
  Jackson, Le, Li, Lin, Lin, Ma, Maddukuri, Mirchandani, Morton, Nguyen,
  O'Neill, Scalise, Seale, Son, Tian, Tran, Wang, Wu, Xie, Yang, Yin, Zhang,
  Bastani, Berseth, Bohg, Goldberg, Gupta, Gupta, Jayaraman, Lim, Malik,
  Martín-Martín, Ramamoorthy, Sadigh, Song, Wu, Yip, Zhu, Kollar, Levine, and
  Finn]{khazatsky2024droid}
Alexander Khazatsky, Karl Pertsch, Suraj Nair, Ashwin Balakrishna, Sudeep
  Dasari, Siddharth Karamcheti, Soroush Nasiriany, Mohan~Kumar Srirama,
  Lawrence~Yunliang Chen, Kirsty Ellis, Peter~David Fagan, Joey Hejna, Masha
  Itkina, Marion Lepert, Yecheng~Jason Ma, Patrick~Tree Miller, Jimmy Wu,
  Suneel Belkhale, Shivin Dass, Huy Ha, Arhan Jain, Abraham Lee, Youngwoon Lee,
  Marius Memmel, Sungjae Park, Ilija Radosavovic, Kaiyuan Wang, Albert Zhan,
  Kevin Black, Cheng Chi, Kyle~Beltran Hatch, Shan Lin, Jingpei Lu, Jean
  Mercat, Abdul Rehman, Pannag~R Sanketi, Archit Sharma, Cody Simpson, Quan
  Vuong, Homer~Rich Walke, Blake Wulfe, Ted Xiao, Jonathan~Heewon Yang, Arefeh
  Yavary, Tony~Z. Zhao, Christopher Agia, Rohan Baijal, Mateo~Guaman Castro,
  Daphne Chen, Qiuyu Chen, Trinity Chung, Jaimyn Drake, Ethan~Paul Foster,
  Jensen Gao, David~Antonio Herrera, Minho Heo, Kyle Hsu, Jiaheng Hu, Donovon
  Jackson, Charlotte Le, Yunshuang Li, Kevin Lin, Roy Lin, Zehan Ma, Abhiram
  Maddukuri, Suvir Mirchandani, Daniel Morton, Tony Nguyen, Abigail O'Neill,
  Rosario Scalise, Derick Seale, Victor Son, Stephen Tian, Emi Tran, Andrew~E.
  Wang, Yilin Wu, Annie Xie, Jingyun Yang, Patrick Yin, Yunchu Zhang, Osbert
  Bastani, Glen Berseth, Jeannette Bohg, Ken Goldberg, Abhinav Gupta, Abhishek
  Gupta, Dinesh Jayaraman, Joseph~J Lim, Jitendra Malik, Roberto
  Martín-Martín, Subramanian Ramamoorthy, Dorsa Sadigh, Shuran Song, Jiajun
  Wu, Michael~C. Yip, Yuke Zhu, Thomas Kollar, Sergey Levine, and Chelsea Finn.
\newblock Droid: A large-scale in-the-wild robot manipulation dataset.
\newblock In \emph{Proceedings of Robotics: Science and Systems}, 2024.

\bibitem[Kim et~al.(2024)Kim, Pertsch, Karamcheti, Xiao, Balakrishna, Nair,
  Rafailov, Foster, Lam, Sanketi, et~al.]{kim2024openvla}
Moo~Jin Kim, Karl Pertsch, Siddharth Karamcheti, Ted Xiao, Ashwin Balakrishna,
  Suraj Nair, Rafael Rafailov, Ethan Foster, Grace Lam, Pannag Sanketi, et~al.
\newblock Openvla: An open-source vision-language-action model.
\newblock \emph{arXiv preprint arXiv:2406.09246}, 2024.

\bibitem[Kim et~al.(2025)Kim, Finn, and Liang]{kim2025fine}
Moo~Jin Kim, Chelsea Finn, and Percy Liang.
\newblock Fine-tuning vision-language-action models: Optimizing speed and
  success.
\newblock \emph{arXiv preprint arXiv:2502.19645}, 2025.

\bibitem[Laurençon et~al.(2023)Laurençon, Saulnier, Tronchon, Bekman, Singh,
  Lozhkov, Wang, Karamcheti, Rush, Kiela, Cord, and Sanh]{laurencon2023obelics}
Hugo Laurençon, Lucile Saulnier, Léo Tronchon, Stas Bekman, Amanpreet Singh,
  Anton Lozhkov, Thomas Wang, Siddharth Karamcheti, Alexander~M. Rush, Douwe
  Kiela, Matthieu Cord, and Victor Sanh.
\newblock {OBELICS}: An open web-scale filtered dataset of interleaved
  image-text documents.
\newblock In \emph{Neural Information Processing Systems Track on Datasets and
  Benchmarks (NeurIPS Datasets and Benchmarks)}, 2023.

\bibitem[Li et~al.(2024)Li, Liang, Wang, Luo, Chen, Liao, Wei, Deng, Xu, Zhang,
  et~al.]{li2024cogact}
Qixiu Li, Yaobo Liang, Zeyu Wang, Lin Luo, Xi~Chen, Mozheng Liao, Fangyun Wei,
  Yu~Deng, Sicheng Xu, Yizhong Zhang, et~al.
\newblock Cogact: A foundational vision-language-action model for synergizing
  cognition and action in robotic manipulation.
\newblock \emph{arXiv preprint arXiv:2411.19650}, 2024.

\bibitem[Liang et~al.(2024)Liang, Yu, Luo, Iyer, Dong, Zhou, Ghosh, Lewis, Yih,
  Zettlemoyer, et~al.]{liang2024mixture}
Weixin Liang, Lili Yu, Liang Luo, Srinivasan Iyer, Ning Dong, Chunting Zhou,
  Gargi Ghosh, Mike Lewis, Wen-tau Yih, Luke Zettlemoyer, et~al.
\newblock Mixture-of-transformers: A sparse and scalable architecture for
  multi-modal foundation models.
\newblock \emph{arXiv preprint arXiv:2411.04996}, 2024.

\bibitem[Lipman et~al.(2022)Lipman, Chen, Ben-Hamu, Nickel, and
  Le]{lipman2022flow}
Yaron Lipman, Ricky~TQ Chen, Heli Ben-Hamu, Maximilian Nickel, and Matt Le.
\newblock Flow matching for generative modeling.
\newblock \emph{arXiv preprint arXiv:2210.02747}, 2022.

\bibitem[Liu et~al.(2024{\natexlab{a}})Liu, Zhu, Gao, Feng, Liu, Zhu, and
  Stone]{liu2024libero}
Bo~Liu, Yifeng Zhu, Chongkai Gao, Yihao Feng, Qiang Liu, Yuke Zhu, and Peter
  Stone.
\newblock Libero: Benchmarking knowledge transfer for lifelong robot learning.
\newblock \emph{Advances in Neural Information Processing Systems}, 36,
  2024{\natexlab{a}}.

\bibitem[Liu et~al.(2023)Liu, Li, Wu, and Lee]{liu2023llava}
Haotian Liu, Chunyuan Li, Qingyang Wu, and Yong~Jae Lee.
\newblock Visual instruction tuning.
\newblock In \emph{Advances in Neural Information Processing Systems
  (NeurIPS)}, 2023.

\bibitem[Liu et~al.(2025)Liu, Chen, An, Liu, Zhang, Gu, Li, Guo, Chen, Liu,
  et~al.]{liu2025hybridvla}
Jiaming Liu, Hao Chen, Pengju An, Zhuoyang Liu, Renrui Zhang, Chenyang Gu,
  Xiaoqi Li, Ziyu Guo, Sixiang Chen, Mengzhen Liu, et~al.
\newblock Hybridvla: Collaborative diffusion and autoregression in a unified
  vision-language-action model.
\newblock \emph{arXiv preprint arXiv:2503.10631}, 2025.

\bibitem[Liu(2022)]{liu2022rectified}
Qiang Liu.
\newblock Rectified flow: A marginal preserving approach to optimal transport.
\newblock \emph{arXiv preprint arXiv:2209.14577}, 2022.

\bibitem[Liu et~al.(2024{\natexlab{b}})Liu, Wu, Li, Tan, Chen, Wang, Xu, Su,
  and Zhu]{liu2024rdt}
Songming Liu, Lingxuan Wu, Bangguo Li, Hengkai Tan, Huayu Chen, Zhengyi Wang,
  Ke~Xu, Hang Su, and Jun Zhu.
\newblock Rdt-1b: a diffusion foundation model for bimanual manipulation.
\newblock \emph{arXiv preprint arXiv:2410.07864}, 2024{\natexlab{b}}.

\bibitem[Liu et~al.(2022)Liu, Gong, and Liu]{liu2022flow}
Xingchao Liu, Chengyue Gong, and Qiang Liu.
\newblock Flow straight and fast: Learning to generate and transfer data with
  rectified flow.
\newblock \emph{arXiv preprint arXiv:2209.03003}, 2022.

\bibitem[{Open X-Embodiment Collaboration} et~al.(2023){Open X-Embodiment
  Collaboration}, Padalkar, Pooley, Jain, Bewley, Herzog, Irpan, Khazatsky,
  Rai, Singh, Brohan, Raffin, Wahid, Burgess-Limerick, Kim, Schölkopf, Ichter,
  Lu, Xu, Finn, Xu, Chi, Huang, Chan, Pan, Fu, Devin, Driess, Pathak, Shah,
  Büchler, Kalashnikov, Sadigh, Johns, Ceola, Xia, Stulp, Zhou, Sukhatme,
  Salhotra, Yan, Schiavi, Su, Fang, Shi, Amor, Christensen, Furuta, Walke,
  Fang, Mordatch, Radosavovic, Leal, Liang, Kim, Schneider, Hsu, Bohg, Bingham,
  Wu, Wu, Luo, Gu, Tan, Oh, Malik, Tompson, Yang, Lim, Silvério, Han, Rao,
  Pertsch, Hausman, Go, Gopalakrishnan, Goldberg, Byrne, Oslund, Kawaharazuka,
  Zhang, Majd, Rana, Srinivasan, Chen, Pinto, Tan, Ott, Lee, Tomizuka, Du, Ahn,
  Zhang, Ding, Srirama, Sharma, Kim, Kanazawa, Hansen, Heess, Joshi,
  Suenderhauf, Palo, Shafiullah, Mees, Kroemer, Sanketi, Wohlhart, Xu,
  Sermanet, Sundaresan, Vuong, Rafailov, Tian, Doshi, Martín-Martín,
  Mendonca, Shah, Hoque, Julian, Bustamante, Kirmani, Levine, Moore, Bahl,
  Dass, Song, Xu, Haldar, Adebola, Guist, Nasiriany, Schaal, Welker, Tian,
  Dasari, Belkhale, Osa, Harada, Matsushima, Xiao, Yu, Ding, Davchev, Zhao,
  Armstrong, Darrell, Jain, Vanhoucke, Zhan, Zhou, Burgard, Chen, Wang, Zhu,
  Li, Lu, Chebotar, Zhou, Zhu, Xu, Wang, Bisk, Cho, Lee, Cui, hua Wu, Tang,
  Zhu, Li, Iwasawa, Matsuo, Xu, and Cui]{open_x_embodiment_rt_x_2023}
{Open X-Embodiment Collaboration}, Abhishek Padalkar, Acorn Pooley, Ajinkya
  Jain, Alex Bewley, Alex Herzog, Alex Irpan, Alexander Khazatsky, Anant Rai,
  Anikait Singh, Anthony Brohan, Antonin Raffin, Ayzaan Wahid, Ben
  Burgess-Limerick, Beomjoon Kim, Bernhard Schölkopf, Brian Ichter, Cewu Lu,
  Charles Xu, Chelsea Finn, Chenfeng Xu, Cheng Chi, Chenguang Huang, Christine
  Chan, Chuer Pan, Chuyuan Fu, Coline Devin, Danny Driess, Deepak Pathak, Dhruv
  Shah, Dieter Büchler, Dmitry Kalashnikov, Dorsa Sadigh, Edward Johns,
  Federico Ceola, Fei Xia, Freek Stulp, Gaoyue Zhou, Gaurav~S. Sukhatme, Gautam
  Salhotra, Ge~Yan, Giulio Schiavi, Hao Su, Hao-Shu Fang, Haochen Shi, Heni~Ben
  Amor, Henrik~I Christensen, Hiroki Furuta, Homer Walke, Hongjie Fang, Igor
  Mordatch, Ilija Radosavovic, Isabel Leal, Jacky Liang, Jaehyung Kim, Jan
  Schneider, Jasmine Hsu, Jeannette Bohg, Jeffrey Bingham, Jiajun Wu, Jialin
  Wu, Jianlan Luo, Jiayuan Gu, Jie Tan, Jihoon Oh, Jitendra Malik, Jonathan
  Tompson, Jonathan Yang, Joseph~J. Lim, João Silvério, Junhyek Han, Kanishka
  Rao, Karl Pertsch, Karol Hausman, Keegan Go, Keerthana Gopalakrishnan, Ken
  Goldberg, Kendra Byrne, Kenneth Oslund, Kento Kawaharazuka, Kevin Zhang,
  Keyvan Majd, Krishan Rana, Krishnan Srinivasan, Lawrence~Yunliang Chen,
  Lerrel Pinto, Liam Tan, Lionel Ott, Lisa Lee, Masayoshi Tomizuka, Maximilian
  Du, Michael Ahn, Mingtong Zhang, Mingyu Ding, Mohan~Kumar Srirama, Mohit
  Sharma, Moo~Jin Kim, Naoaki Kanazawa, Nicklas Hansen, Nicolas Heess, Nikhil~J
  Joshi, Niko Suenderhauf, Norman~Di Palo, Nur Muhammad~Mahi Shafiullah, Oier
  Mees, Oliver Kroemer, Pannag~R Sanketi, Paul Wohlhart, Peng Xu, Pierre
  Sermanet, Priya Sundaresan, Quan Vuong, Rafael Rafailov, Ran Tian, Ria Doshi,
  Roberto Martín-Martín, Russell Mendonca, Rutav Shah, Ryan Hoque, Ryan
  Julian, Samuel Bustamante, Sean Kirmani, Sergey Levine, Sherry Moore, Shikhar
  Bahl, Shivin Dass, Shuran Song, Sichun Xu, Siddhant Haldar, Simeon Adebola,
  Simon Guist, Soroush Nasiriany, Stefan Schaal, Stefan Welker, Stephen Tian,
  Sudeep Dasari, Suneel Belkhale, Takayuki Osa, Tatsuya Harada, Tatsuya
  Matsushima, Ted Xiao, Tianhe Yu, Tianli Ding, Todor Davchev, Tony~Z. Zhao,
  Travis Armstrong, Trevor Darrell, Vidhi Jain, Vincent Vanhoucke, Wei Zhan,
  Wenxuan Zhou, Wolfram Burgard, Xi~Chen, Xiaolong Wang, Xinghao Zhu, Xuanlin
  Li, Yao Lu, Yevgen Chebotar, Yifan Zhou, Yifeng Zhu, Ying Xu, Yixuan Wang,
  Yonatan Bisk, Yoonyoung Cho, Youngwoon Lee, Yuchen Cui, Yueh hua Wu, Yujin
  Tang, Yuke Zhu, Yunzhu Li, Yusuke Iwasawa, Yutaka Matsuo, Zhuo Xu, and
  Zichen~Jeff Cui.
\newblock Open {X-E}mbodiment: Robotic learning datasets and {RT-X} models.
\newblock \url{https://arxiv.org/abs/2310.08864}, 2023.

\bibitem[Pertsch et~al.(2025)Pertsch, Stachowicz, Ichter, Driess, Nair, Vuong,
  Mees, Finn, and Levine]{pertsch2025fast}
Karl Pertsch, Kyle Stachowicz, Brian Ichter, Danny Driess, Suraj Nair, Quan
  Vuong, Oier Mees, Chelsea Finn, and Sergey Levine.
\newblock {FAST}: Efficient action tokenization for vision-language-action
  models.
\newblock \emph{Robotics: Science and Systems}, 2025.

\bibitem[Reuss et~al.(2024)Reuss, Pari, Agrawal, and
  Lioutikov]{reuss2024efficient}
Moritz Reuss, Jyothish Pari, Pulkit Agrawal, and Rudolf Lioutikov.
\newblock Efficient diffusion transformer policies with mixture of expert
  denoisers for multitask learning.
\newblock \emph{arXiv preprint arXiv:2412.12953}, 2024.

\bibitem[Shafiullah et~al.(2023)Shafiullah, Rai, Etukuru, Liu, Misra, Chintala,
  and Pinto]{shafiullah2023bringingrobotshome}
Nur Muhammad~Mahi Shafiullah, Anant Rai, Haritheja Etukuru, Yiqian Liu, Ishan
  Misra, Soumith Chintala, and Lerrel Pinto.
\newblock On bringing robots home.
\newblock \emph{arXiv preprint arXiv:2311.16098}, 2023.

\bibitem[Shi et~al.(2024)Shi, Han, Zhou, and Liang]{shi2024xi}
Weijia Shi, Xiaochuang Han, Chunting Zhou, and Weixin Liang.
\newblock Xi victoria lin, luke zettlemoyer, and lili yu. llamafusion: Adapting
  pretrained language models for multimodal generation.
\newblock \emph{arXiv preprint arXiv:2412.15188}, 2024.

\bibitem[Szot et~al.(2024)Szot, Mazoure, Attia, Timofeev, Agrawal, Hjelm, Gan,
  Kira, and Toshev]{szot2024multimodal}
Andrew Szot, Bogdan Mazoure, Omar Attia, Aleksei Timofeev, Harsh Agrawal, Devon
  Hjelm, Zhe Gan, Zsolt Kira, and Alexander Toshev.
\newblock From multimodal llms to generalist embodied agents: Methods and
  lessons.
\newblock \emph{arXiv preprint arXiv:2412.08442}, 2024.

\bibitem[Team(2024)]{team2024chameleon}
Chameleon Team.
\newblock Chameleon: Mixed-modal early-fusion foundation models.
\newblock \emph{arXiv preprint arXiv:2405.09818}, 2024.

\bibitem[Team et~al.(2025)Team, Abeyruwan, Ainslie, Alayrac, Arenas, Armstrong,
  Balakrishna, Baruch, Bauza, Blokzijl, Bohez, Bousmalis, Brohan, Buschmann,
  Byravan, Cabi, Caluwaerts, Casarini, Chang, Chen, Chen, Chiang, Choromanski,
  D'Ambrosio, Dasari, Davchev, Devin, Palo, Ding, Dostmohamed, Driess, Du,
  Dwibedi, Elabd, Fantacci, Fong, Frey, Fu, Giustina, Gopalakrishnan, Graesser,
  Hasenclever, Heess, Hernaez, Herzog, Hofer, Humplik, Iscen, Jacob, Jain,
  Julian, Kalashnikov, Karagozler, Karp, Kew, Kirkland, Kirmani, Kuang, Lampe,
  Laurens, Leal, Lee, Lee, Liang, Lin, Maddineni, Majumdar, Michaely, Moreno,
  Neunert, Nori, Parada, Parisotto, Pastor, Pooley, Rao, Reymann, Sadigh,
  Saliceti, Sanketi, Sermanet, Shah, Sharma, Shea, Shu, Sindhwani, Singh,
  Soricut, Springenberg, Sterneck, Surdulescu, Tan, Tompson, Vanhoucke, Varley,
  Vesom, Vezzani, Vinyals, Wahid, Welker, Wohlhart, Xia, Xiao, Xie, Xie, Xu,
  Xu, Xu, Xu, Yang, Yao, Yaroshenko, Yu, Yuan, Zhang, Zhang, Zhou, and
  Zhou]{geminirobotics2025}
Gemini~Robotics Team, Saminda Abeyruwan, Joshua Ainslie, Jean-Baptiste Alayrac,
  Montserrat~Gonzalez Arenas, Travis Armstrong, Ashwin Balakrishna, Robert
  Baruch, Maria Bauza, Michiel Blokzijl, Steven Bohez, Konstantinos Bousmalis,
  Anthony Brohan, Thomas Buschmann, Arunkumar Byravan, Serkan Cabi, Ken
  Caluwaerts, Federico Casarini, Oscar Chang, Jose~Enrique Chen, Xi~Chen,
  Hao-Tien~Lewis Chiang, Krzysztof Choromanski, David D'Ambrosio, Sudeep
  Dasari, Todor Davchev, Coline Devin, Norman~Di Palo, Tianli Ding, Adil
  Dostmohamed, Danny Driess, Yilun Du, Debidatta Dwibedi, Michael Elabd,
  Claudio Fantacci, Cody Fong, Erik Frey, Chuyuan Fu, Marissa Giustina,
  Keerthana Gopalakrishnan, Laura Graesser, Leonard Hasenclever, Nicolas Heess,
  Brandon Hernaez, Alexander Herzog, R.~Alex Hofer, Jan Humplik, Atil Iscen,
  Mithun~George Jacob, Deepali Jain, Ryan Julian, Dmitry Kalashnikov, M.~Emre
  Karagozler, Stefani Karp, Chase Kew, Jerad Kirkland, Sean Kirmani, Yuheng
  Kuang, Thomas Lampe, Antoine Laurens, Isabel Leal, Alex~X. Lee,
  Tsang-Wei~Edward Lee, Jacky Liang, Yixin Lin, Sharath Maddineni, Anirudha
  Majumdar, Assaf~Hurwitz Michaely, Robert Moreno, Michael Neunert, Francesco
  Nori, Carolina Parada, Emilio Parisotto, Peter Pastor, Acorn Pooley, Kanishka
  Rao, Krista Reymann, Dorsa Sadigh, Stefano Saliceti, Pannag Sanketi, Pierre
  Sermanet, Dhruv Shah, Mohit Sharma, Kathryn Shea, Charles Shu, Vikas
  Sindhwani, Sumeet Singh, Radu Soricut, Jost~Tobias Springenberg, Rachel
  Sterneck, Razvan Surdulescu, Jie Tan, Jonathan Tompson, Vincent Vanhoucke,
  Jake Varley, Grace Vesom, Giulia Vezzani, Oriol Vinyals, Ayzaan Wahid, Stefan
  Welker, Paul Wohlhart, Fei Xia, Ted Xiao, Annie Xie, Jinyu Xie, Peng Xu,
  Sichun Xu, Ying Xu, Zhuo Xu, Yuxiang Yang, Rui Yao, Sergey Yaroshenko, Wenhao
  Yu, Wentao Yuan, Jingwei Zhang, Tingnan Zhang, Allan Zhou, and Yuxiang Zhou.
\newblock Gemini robotics: Bringing ai into the physical world, 2025.
\newblock URL \url{https://arxiv.org/abs/2503.20020}.

\bibitem[Team et~al.(2024{\natexlab{a}})Team, Mesnard, Hardin, Dadashi,
  Bhupatiraju, Pathak, Sifre, Rivi{\`e}re, Kale, Love, et~al.]{team2024gemma}
Gemma Team, Thomas Mesnard, Cassidy Hardin, Robert Dadashi, Surya Bhupatiraju,
  Shreya Pathak, Laurent Sifre, Morgane Rivi{\`e}re, Mihir~Sanjay Kale,
  Juliette Love, et~al.
\newblock Gemma: Open models based on gemini research and technology.
\newblock \emph{arXiv preprint arXiv:2403.08295}, 2024{\natexlab{a}}.

\bibitem[Team et~al.(2024{\natexlab{b}})Team, Ghosh, Walke, Pertsch, Black,
  Mees, Dasari, Hejna, Kreiman, Xu, et~al.]{team2024octo}
Octo~Model Team, Dibya Ghosh, Homer Walke, Karl Pertsch, Kevin Black, Oier
  Mees, Sudeep Dasari, Joey Hejna, Tobias Kreiman, Charles Xu, et~al.
\newblock Octo: An open-source generalist robot policy.
\newblock \emph{arXiv preprint arXiv:2405.12213}, 2024{\natexlab{b}}.

\bibitem[Tong et~al.(2024)Tong, Brown, Wu, Woo, IYER, Akula, Yang, Yang,
  Middepogu, Wang, et~al.]{tong2024cambrian}
Peter Tong, Ellis Brown, Penghao Wu, Sanghyun Woo, Adithya Jairam~Vedagiri
  IYER, Sai~Charitha Akula, Shusheng Yang, Jihan Yang, Manoj Middepogu, Ziteng
  Wang, et~al.
\newblock Cambrian-1: A fully open, vision-centric exploration of multimodal
  llms.
\newblock \emph{Advances in Neural Information Processing Systems},
  37:\penalty0 87310--87356, 2024.

\bibitem[Vaswani et~al.(2017)Vaswani, Shazeer, Parmar, Uszkoreit, Jones, Gomez,
  Kaiser, and Polosukhin]{vaswani2017attention}
Ashish Vaswani, Noam Shazeer, Niki Parmar, Jakob Uszkoreit, Llion Jones,
  Aidan~N Gomez, \L~ukasz Kaiser, and Illia Polosukhin.
\newblock Attention is all you need.
\newblock In \emph{Advances in Neural Information Processing Systems},
  volume~30, 2017.

\bibitem[Walke et~al.(2023)Walke, Black, Zhao, Vuong, Zheng, Hansen-Estruch,
  He, Myers, Kim, Du, et~al.]{walke2023bridgedata}
Homer~Rich Walke, Kevin Black, Tony~Z Zhao, Quan Vuong, Chongyi Zheng, Philippe
  Hansen-Estruch, Andre~Wang He, Vivek Myers, Moo~Jin Kim, Max Du, et~al.
\newblock {BridgeData} v2: A dataset for robot learning at scale.
\newblock In \emph{Conference on Robot Learning}, pages 1723--1736. PMLR, 2023.

\bibitem[Wang et~al.(2024)Wang, Lv, Yu, Hong, Qi, Wang, Ji, Yang, Zhao, Song,
  Xu, Xu, Li, Dong, Ding, and Tang]{cogvlm}
Weihan Wang, Qingsong Lv, Wenmeng Yu, Wenyi Hong, Ji~Qi, Yan Wang, Junhui Ji,
  Zhuoyi Yang, Lei Zhao, Xixuan Song, Jiazheng Xu, Bin Xu, Juanzi Li, Yuxiao
  Dong, Ming Ding, and Jie Tang.
\newblock Cogvlm: Visual expert for pretrained language models, 2024.
\newblock URL \url{https://arxiv.org/abs/2311.03079}.

\bibitem[Wen et~al.(2024)Wen, Zhu, Li, Zhu, Wu, Xu, Liu, Cheng, Shen, Peng,
  Feng, and Tang]{wen2024tinyvlafastdataefficientvisionlanguageaction}
Junjie Wen, Yichen Zhu, Jinming Li, Minjie Zhu, Kun Wu, Zhiyuan Xu, Ning Liu,
  Ran Cheng, Chaomin Shen, Yaxin Peng, Feifei Feng, and Jian Tang.
\newblock Tinyvla: Towards fast, data-efficient vision-language-action models
  for robotic manipulation.
\newblock \emph{arXiv preprint arXiv:2409.12514}, 2024.

\bibitem[Wen et~al.(2025)Wen, Zhu, Li, Tang, Shen, and Feng]{wen2025dexvla}
Junjie Wen, Yichen Zhu, Jinming Li, Zhibin Tang, Chaomin Shen, and Feifei Feng.
\newblock Dexvla: Vision-language model with plug-in diffusion expert for
  general robot control.
\newblock \emph{arXiv preprint arXiv:2502.05855}, 2025.

\bibitem[Yu et~al.(2024)Yu, Sun, Zhang, Cui, Zhang, Cao, Wang, and
  Liu]{yu2024capsfusion}
Qiying Yu, Quan Sun, Xiaosong Zhang, Yufeng Cui, Fan Zhang, Yue Cao, Xinlong
  Wang, and Jingjing Liu.
\newblock Capsfusion: Rethinking image-text data at scale.
\newblock In \emph{Proceedings of the IEEE/CVF Conference on Computer Vision
  and Pattern Recognition}, pages 14022--14032, 2024.

\bibitem[Zawalski et~al.(2024)Zawalski, Chen, Pertsch, Mees, Finn, and
  Levine]{Zawalski24-ecot}
Michał Zawalski, William Chen, Karl Pertsch, Oier Mees, Chelsea Finn, and
  Sergey Levine.
\newblock Robotic control via embodied chain-of-thought reasoning.
\newblock In \emph{Conference on Robot Learning}, 2024.

\bibitem[Zhao et~al.(2023)Zhao, Kumar, Levine, and Finn]{zhao2023learning}
Tony~Z Zhao, Vikash Kumar, Sergey Levine, and Chelsea Finn.
\newblock Learning fine-grained bimanual manipulation with low-cost hardware.
\newblock \emph{arXiv preprint arXiv:2304.13705}, 2023.

\bibitem[Zhao et~al.(2024)Zhao, Tompson, Driess, Florence, Ghasemipour, Finn,
  and Wahid]{zhao2024alohaunleashedsimplerecipe}
Tony~Z Zhao, Jonathan Tompson, Danny Driess, Pete Florence, Kamyar Ghasemipour,
  Chelsea Finn, and Ayzaan Wahid.
\newblock Aloha unleashed: A simple recipe for robot dexterity.
\newblock \emph{arXiv preprint arXiv:2410.13126}, 2024.

\bibitem[Zhen et~al.(2024)Zhen, Qiu, Chen, Yang, Yan, Du, Hong, and
  Gan]{zhen20243dvla}
Haoyu Zhen, Xiaowen Qiu, Peihao Chen, Jincheng Yang, Xin Yan, Yilun Du, Yining
  Hong, and Chuang Gan.
\newblock 3d-vla: 3d vision-language-action generative world model.
\newblock \emph{arXiv preprint arXiv:2403.09631}, 2024.

\bibitem[Zheng et~al.(2024)Zheng, Cheng, Daum{\'e}~III, Huang, and
  Kolobov]{zheng2024prise}
Ruijie Zheng, Ching-An Cheng, Hal Daum{\'e}~III, Furong Huang, and Andrey
  Kolobov.
\newblock Prise: Learning temporal action abstractions as a sequence
  compression problem, 2024.

\bibitem[Zhou et~al.(2024)Zhou, Yu, Babu, Tirumala, Yasunaga, Shamis, Kahn, Ma,
  Zettlemoyer, and Levy]{zhou2024transfusion}
Chunting Zhou, Lili Yu, Arun Babu, Kushal Tirumala, Michihiro Yasunaga, Leonid
  Shamis, Jacob Kahn, Xuezhe Ma, Luke Zettlemoyer, and Omer Levy.
\newblock Transfusion: Predict the next token and diffuse images with one
  multi-modal model.
\newblock \emph{arXiv preprint arXiv:2408.11039}, 2024.

\bibitem[Zitkovich et~al.(2023)Zitkovich, Yu, Xu, Xu, Xiao, Xia, Wu, Wohlhart,
  Welker, Wahid, et~al.]{rt22023arxiv}
Brianna Zitkovich, Tianhe Yu, Sichun Xu, Peng Xu, Ted Xiao, Fei Xia, Jialin Wu,
  Paul Wohlhart, Stefan Welker, Ayzaan Wahid, et~al.
\newblock Rt-2: Vision-language-action models transfer web knowledge to robotic
  control.
\newblock In \emph{Conference on Robot Learning}, pages 2165--2183. PMLR, 2023.

\end{thebibliography}

\appendix

%%%%%%%%%%%%%%%%%%%%%%%%%%%%%%%%%%%%%%%%%%%%%%%%%%%%%%%%%%%%

\newpage

\begin{figure}
    \centering
    \includegraphics[width=0.5\textwidth]{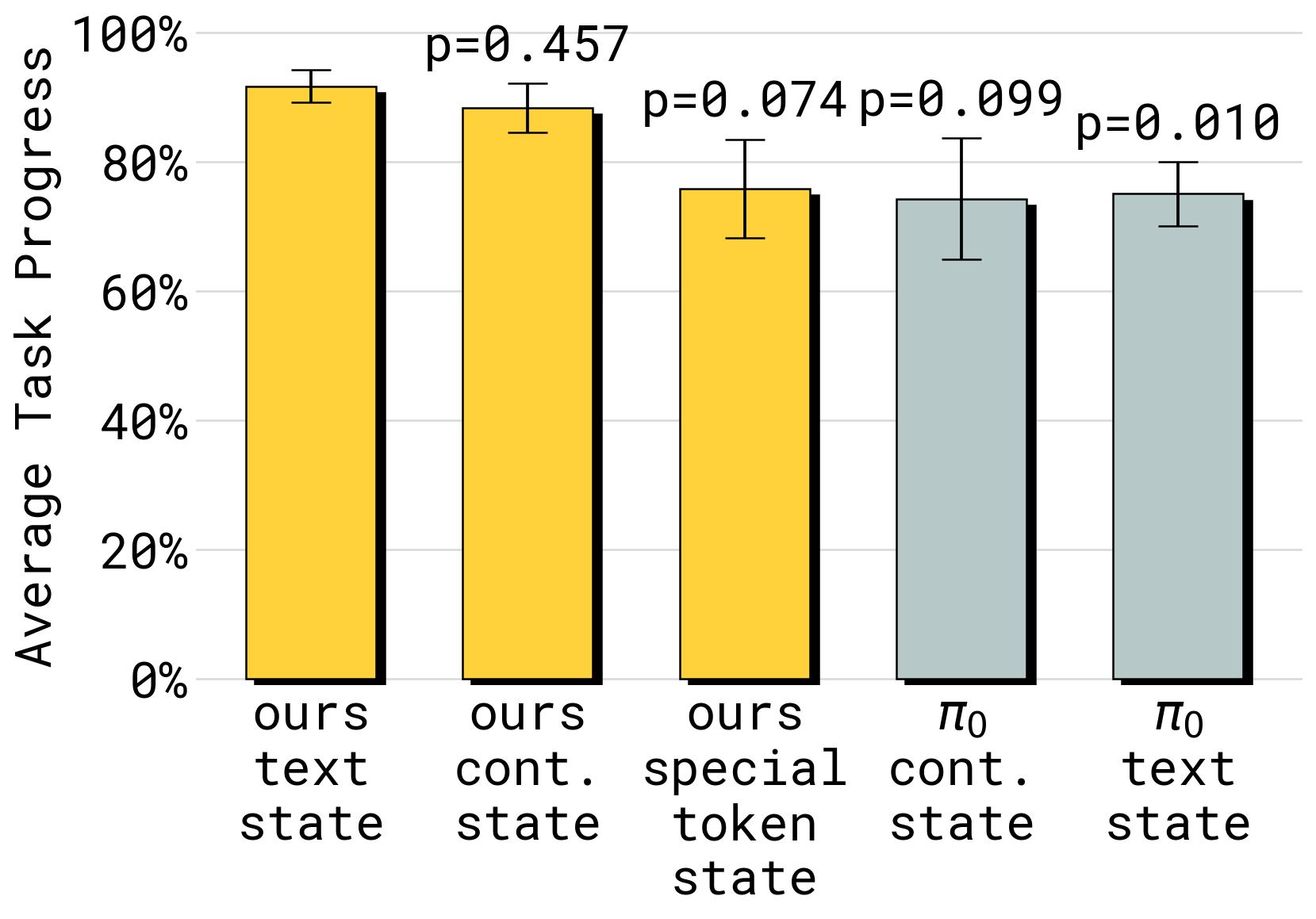}
    \setlength{\abovecaptionskip}{0pt}
    \caption{\small Comparison of different state representations on ``table bussing'' task. Our method works well with both text and continuous state, while $\pi_0$ works worse with both state representations. Therefore, the difference between $\pi_0$ and our method cannot be explained by the difference in their original state representations. Using special tokens as state performs worse.}
    \vspace{5mm}
    \label{fig:stateRepresentations}
\end{figure}

\section{Dataset \& task details}\label{app:dataAndTasks}

\subsection{Common public benchmarks}
We evaluate our method on the simulated LIBERO \cite{liu2024libero} benchmark, as well as the real world DROID \cite{khazatsky2024droid} benchmark.
For these two benchmarks, we evaluate specialist models that are trained only on the respective datasets. For LIBERO, we also evaluate a model fine-tuned from our generalist model that is trained with diverse robot embodiment data and non-robot data. The generalist model did not include LIBERO data during its original training.

\paragraph{DROID.} We evaluate on the same set of real-world tabletop manipulation tasks as \citet{pertsch2025fast}. These tasks include picking and placing, wiping, and opening and closing drawers. The environment and objects are completely unseen. Each trial is scored based on task progress, and we report the average score. 

\paragraph{LIBERO.} The LIBERO \cite{liu2024libero} simulation benchmark consists of four task suites: LIBERO-Spatial, LIBERO-Object, LIBERO-Goal and LIBERO-100 (which is further split into LIBERO-90 and LIBERO-10, also known as LIBERO-Long). We evaluate LIBERO-Spatial, LIBERO-Object, LIBERO-Goal and LIBERO-Long following the same setup as \citet{pertsch2025fast}, jointly training one policy on all four datasets. We additionally evaluate our method on LIBERO-90 with a policy trained only on that dataset. All training datasets are prepared in the same manner as \citet{kim2024openvla}, which includes re-rendering images at a higher resolution of 224 $\times$ 224 px, and filtering out unsuccessful demonstrations and ``no-op'' actions. All methods in our comparisons use both third-person and wrist camera images as inputs.

\subsection{Real-world tasks}
For all our real world tasks, we evaluate 10 episodes per task per policy and report performance as outlined below.
We report statistical significance according to a two-sided t-test.

\subsubsection{Tasks with static robots}
\paragraph{Items in drawer.} The task begins with three objects on a countertop. The static single-arm robot must open the drawer beneath the counter, place the items into the drawer, and close the drawer. One point is awarded for (1) opening the drawer, (2) putting one item into the drawer, and (3) closing the drawer. Thus the maximum number of points is 5. We evaluate a specialist model with data only from static single-arm robots.
This dataset contains more than just this task.
The evaluation environment for this task is unseen.

\paragraph{T-shirt folding.} The task begins with a shirt flat on the table. The static bimanual robot must fold it into the usual folded form with collar facing up. The maximum number of points is 5, and it is scored based on the squareness and amount of wrinkles of the final form. We evaluate a specialist model with data only from static bimanual robots. The dataset contains many different tasks.

\paragraph{Table bussing.} The task begins with 12 objects on the table and two receptacles, one for utensils/dishes and one for trash. The static single-arm robot must follow language commands to pick up the correct object and place it into the correct receptacles. One point is awarded for each correctly placed object. Thus, the maximum number of points is 12. We evaluate a specialist model with data only from static single-arm robots, as well as a generalist model trained with diverse data.

\subsubsection{Tasks with mobile manipulator robots} \label{sec:tasksMobileManipulator}

\paragraph{Make bed.} The task begins with the bed partially unmade, with two gray pillows at the foot of the bed. The mobile bimanual robot must tidy the blanket and place the two pillows at the head of the bed. One point is awarded for (1) straightening the blanket so it covers the sheets, (2) placing one pillow at the head of the bed, (3) blanket being straightened very neatly, and (4) both pillows are placed very neatly. Thus the maximum number of points is 5. We evaluate the generalist model.

\paragraph{Dish in sink.} The task begins with 4 dishes (e.g. plates, bowls, cutting boards, utensils) placed near a sink. The mobile bimanual robot must place all of them in the sink. One point is awarded for (1) picking one item up, and (2) placing one item into the sink. Thus the maximum number of points is 8. We evaluate the generalist model.

\paragraph{Mobile items in drawer.} The task begins with a household item on a countertop. The mobile bimanual robot must place the item into a drawer beneath the counter. One point is awarded for (1) picking up the object, (2) opening the drawer, (3) putting the object into the drawer, (4) closing the drawer.
Therefore, the maximum score 4 points.

\paragraph{Laundry in basket.} The task begins with clothing lying on the ground. The mobile bimanual robot has to pick up the clothing and place it in the laundry basket. One point is awarded for (1) navigating to and picking up the clothing, (2) placing the clothing into or onto the laundry basket, (3) the clothing is fully inside the basket. Therefore, the maximum score is 3 points.

\subsection{Datasets for training the generalist model}

The generalist model is trained on a large dataset encompassing 12 configurations of robot embodiments, including single-arm static manipulators (ARX, UR5, Franka), bimanual static manipulators (ARX, AgileX, Trossen, UR5), and bimanual mobile manipulators (mobile Trossen, ARX slate, Galaxea G1, Hexmove H1, Fibocom). This robot data includes a large diversity of tasks, going much beyond the evaluation tasks considered in this work, e.g., grinding coffee beans or hanging a towel on the oven handle) in diverse environments (both office-like ones and real home ones). We also include the open-source OXE dataset \cite{open_x_embodiment_rt_x_2023}. 

We also train the generalist model with a variety of general VLM tasks. The data involves image captioning (CapsFusion \cite{yu2024capsfusion}, COCO \cite{chen2015microsoft}), visual-question-answering (Cambrian-7M \cite{tong2024cambrian}, PixMo \cite{deitke2024molmo}, VQAv2 \cite{goyal2017making}), as well as object localization. For object localization, we further extend the standard datasets with additional web data of indoor scenes and household objects with bounding box annotations.

\section{Training details}\label{app:trainingDetails}

We use the PaliGemma VLM~\citep{beyer2024paligemma} architecture as the VLM backbone and initialize it with its pre-trained weights. The action expert is a smaller transformer that takes in a sequence of noisy actions $a_{1:H}^{\tau,\omega}$ for an action horizon of 50, i.e.\ $H=50$.
The noisy action chunk is first projected to the transformer embedding dimension using a single linear layer. We use a MLP to project $\tau$ and then applies adaptive RMSNorm to inject the timestep information to each layer of the action expert.
The MLP takes in the form of $\mathrm{swish}(W_2 \cdot \mathrm{swish} (W_1 \cdot \phi(\tau)))$, where $\phi : \mathbb{R} \to \mathbb{R}^w$ is a sinusoidal positional encoding function~\citep{vaswani2017attention} and $W_1, W_2 \in \mathbb{R}^{w \times w}$. The action expert outputs action tokens $y^a_{1:H}$, which are then decoded into the target vector field using a final linear projection.

The dimensions of the VLM backbone and action expert are as follows:
\{\textit{width}=2048, \textit{depth}=18, \textit{mlp\_dim}=16,384, \textit{num\_heads}=18, \textit{num\_kv\_heads}=1, \textit{head\_dim}=256\} for the 2B language model backbone, and the same except for \{\textit{width}=1024, \textit{mlp\_dim}=4096\} for the action expert, leading to 300M parameters.

Embeddings from the VLM and action expert interact only through self-attention. A full prefix mask is used on images, language tokens, and text state; FAST action tokens attend to this prefix and auto-regressively on previous action tokens. Embeddings from the action expert attend to the prefix and to one another, but do not attend to FAST action tokens to avoid information leakage between the two representations of actions. In effect, information flows unidirectionally from the VLM to the action expert; no VLM embedding attends to the action expert.

We follow $\pi_0$ for sampling the flow-matching timestep $\tau$. In summary we deviate from standard uniform sampling $\tau \sim \mathcal{U}(0, 1)$~\citep{lipman2022flow,liu2022rectified} or methods emphasizing midrange timesteps~\citep{esser2024scaling}, 
and instead use a time-step sampling distribution that emphasizes low time-steps \citep{black2024pi_0}, given by $p(\tau) = \mathrm{Beta}(\frac{s-\tau}{s}; \alpha=1.5, \beta=1), s=0.999$.

\section{State representations}\label{sec:stateRepresentations}
We consider three different representations for the robot's proprioceptive state $q\in\mathbb{R}^{s}$.
\begin{itemize}[leftmargin=*]
    \item \textbf{Text state.} This representation discretizes the state into $n_b$ many bins, then converting the bins into numbers from $1$ to $n_b$. Those numbers are then input to the model as normal text. This has been the standard approach for previous VLAs. For tokenizers such as in Gemma \cite{team2024gemma} this implies up to $\lfloor\log_{10}(n_b) + 2)\rfloor\cdot s$ many tokens to represent the state as text.
    \item \textbf{Special token state.} Similar to the ``text state'', this representation uses the discretized bins directly, associating each bin with a special token from the VLM tokenizer. This requires $s$ tokens.
    \item \textbf{Continuous state.} This representation inputs the real-valued vector $q$ directly into the model by projecting it with a learned affine projection into the $d_e$ dimensional embedding space.
\end{itemize}
The text state approach has the advantage of it being the closest to what the backbone might have seen during pre-training as it is just a sequence of numbers in natural text but it requires the most tokens. Both special token state and continuous state are completely new inputs to the model with randomly initialized projections/embeddings.

\end{document}